%% file: main.tex
\pdfoutput=1
\documentclass[10pt,twocolumn,letterpaper]{article}

\usepackage[pagenumbers]{cvpr}

\input{preamble}

\usepackage{booktabs}
\usepackage{array}
\usepackage{multirow}
\usepackage{xspace}
\usepackage{amsmath}
\usepackage{amssymb}
\usepackage{algorithm}
\usepackage{algpseudocode}
\usepackage{tikz}
\usetikzlibrary{arrows.meta}
\definecolor{cvprblue}{rgb}{0.21,0.49,0.74}
\definecolor{fmtcol}{HTML}{C2531A}
\usepackage[pagebackref,breaklinks,colorlinks,allcolors=cvprblue]{hyperref}
\hypersetup{pdftitle={FMT: Lossless Image Compression in a Fixed Number of Parallel Steps},pdfauthor={Ammar Ali}}

\usepackage{microtype}
\makeatletter
\renewcommand\paragraph{\@startsection{paragraph}{4}{\z@}{1.3ex \@plus.5ex \@minus.2ex}{-1em}%
  {\normalfont\normalsize\bfseries}}
\makeatother

\newcommand{\fmt}{FMT\xspace}
\newcommand{\bpsp}{bits per subpixel\xspace}
\newcommand{\x}{\ensuremath{\times}}
\newcommand{\best}[1]{\textbf{#1}}

\def\confName{CVPR}
\def\confYear{2027}

\title{FMT: Lossless Image Compression in a Fixed Number of Parallel Steps}

\author{Ammar Ali\\
LEAD Lab, ITMO University}

\begin{document}
\maketitle
\input{fig/teaser}
\input{sec/abstract}
\input{sec/intro}
\input{sec/related_work}
\input{sec/method}
\input{sec/experiments}
\input{sec/conclusion}
{
    \small
    \input{strefs}
    \bibliographystyle{ieeenat_fullname}
    \bibliography{refs}
}

\input{supp/supp}

\end{document}

%% file: preamble.tex
\usepackage{cuted} %

\usepackage{currfile} %

\usepackage{caption} %

%% file: fig/teaser.tex
\begin{strip}
\centering
\vspace{-6mm}
\includegraphics[width=\linewidth]{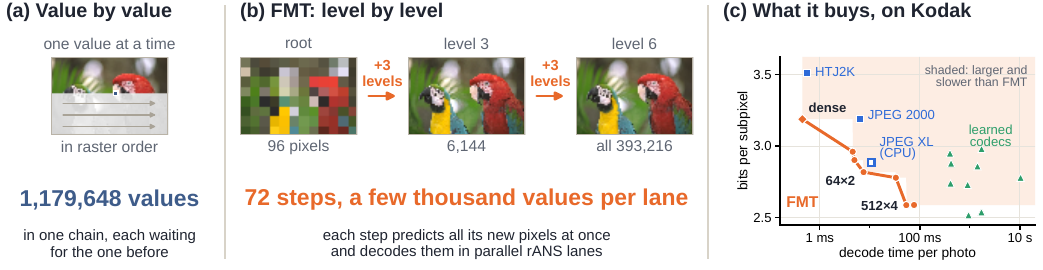}
\vspace{-5mm}
\captionsetup{hypcap=false}
\captionof{figure}{\textbf{Two ways to decode a 768\x512 photo, and what the second buys.}
(a) A sequential decoder, such as the arithmetic decoder of a pixel-by-pixel model or of DLPR~\cite{dlpr}, reads its
1,179,648 values one after another. (b) \fmt stores the photo as a coarse-to-fine tree: a root of one pixel in 64 along each axis,
then levels that add the pixels halfway between the decoded ones, as corrections to predictions from them. A step's
pixels are independent, so a GPU predicts them at once and decodes them in parallel lanes: 72 steps (3 channels \x\ 6
levels \x\ 4 steps), a few thousand values per lane. (c) \fmt on one RTX 4090 against classical codecs (JPEG XL on a
32-thread CPU) and learned codecs at their published times: every codec in the shaded area is larger and slower than
an \fmt mode (all codecs: \cref{fig:kodak}).}
\label{fig:teaser}
\end{strip}

%% file: sec/abstract.tex
\begin{abstract}
Learned lossless codecs make smaller files than common classical codecs, yet the images we archive and train on still use classical codecs. A main reason is time: a learned codec takes seconds to decode a photo that a classical codec opens in milliseconds, and their faster variants give up most of their gain in size. We argue that much of this time is spent waiting, not computing: the network waits for the values it conditions on, and the arithmetic decoder reads them one at a time. We introduce \fmt (Fixed-step Multiscale Tree), a lossless image format whose decoding order and entropy coder are designed together so that neither wait grows with the image. \fmt stores an image as a coarse-to-fine tree whose steps hold independent values: a GPU predicts each step at once and decodes it in parallel rANS lanes that carry their states from step to step, so an image of any size decodes in a fixed number of sequential steps. On this tree we build a family of bit-exact codecs, from a rule with no network to large networks, and no codec we ran on the same GPU beats any of them on both size and speed. Without a network, \fmt beats NVIDIA's HTJ2K on size and speed; with a large one, it decodes Kodak photos 9 to 61\x\ faster than learned codecs we ran on the same A100, at the same size or smaller.

\end{abstract}

%% file: sec/intro.tex
\section{Introduction}
\label{sec:intro}

Many images must be stored without losing a single pixel: originals in archives, frames in editing pipelines, and the training data of models that restore images. Such images are written once and read many times, so how fast they decode matters as much as how small they are. Classical lossless codecs, such as PNG, JPEG-LS, JPEG XL and HTJ2K, predict or transform pixels with hand-designed rules~\cite{calic,locoi,png,jpegxl,htj2k} and decode a photo in milliseconds. Learned codecs replace these rules with neural networks and make clearly smaller files~\cite{l3c,dlpr,aribbps,hpac,callic}, but they take seconds per photo, tens to hundreds of times longer (\cref{fig:kodak}), and their faster designs give up most of the gain in size~\cite{hpac,pilc}. One must still choose between size and speed.

\begin{table}[t]
\centering
\small
\setlength{\tabcolsep}{3.5pt}
\begin{tabular}{@{}lrrl@{}}
\toprule
Decoding order & Steps & Coder chain & A value sees \\
\midrule
Value by value~\cite{pixelcnn} & 1,179,648 & 1,179,648 & above and left \\
JPEG XL groups~\cite{jxloverview} & 196,608 & 196,608 & its block \\
HTJ2K blocks~\cite{htj2k} & 5 & 4,096 & its block \\
Interpolation~\cite{kamisli} & about 15 & 1,179,648 & all sides \\
Patch groups~\cite{hpac} & 94 & -- & earlier groups \\
\fmt tree & 36 or 72 & $\le$ 7,410 & all sides \\
\bottomrule
\end{tabular}
\vspace{-2mm}
\caption{\textbf{Sequential work} of decoding a 768\x512 photo, 1,179,648 values. Steps: runs of the predictor that must follow one another (Kamisli's: 5 scales \x\ 3 interpolators; HTJ2K: 5 wavelet levels). Coder chain: the most values one entropy decoder decodes one after another (JPEG XL, HTJ2K: per $256\times256$ group and $64\times64$ code-block, decoded independently; Kamisli: one CPU coder; HPAC: one bitstream, decoding not described; \fmt: our 512\x4 on Kodak, at most $T=8{,}192$ plus one per step, \cref{sec:decoding}).}
\label{tab:why}
\vspace{-3mm}
\end{table}

Why are learned codecs so slow? We argue that much of their time goes to waiting rather than computing (\cref{tab:why}). An entropy decoder can recover a value only once it knows the value's probability, and a learned codec computes it from values decoded before it. Decoding is therefore a chain with two links. First, the network waits for the values it conditions on: once per value in a pixel-by-pixel model~\cite{pixelcnn}, and still many times in codecs that predict groups of values~\cite{multiscale,srec,lcfdnet,kamisli,hpac}. Second, the arithmetic decoder reads one value after another, often on the CPU. FNLIC~\cite{fnlic} reports 13\,ms of network inference per photo, yet in our runs it took half a second per photo, two fifths of it in such a coder; LC-FDNet~\cite{lcfdnet} took a second, a fifth of it in its coder on the CPU. No number of GPU cores shortens this chain (\cref{fig:teaser}a).

In \fmt, every pixel is the end of a path from a coarse root (\cref{fig:teaser}b), as every weight is in state trees~\cite{statetrees}: each level adds the pixels halfway between those already decoded, predicts them from their decoded neighbours and stores only a correction (the residual), so neighbouring pixels share the top of their paths, the coarser pixels they are predicted from. Each correction is in turn binarised into a short path of decisions: its size, then its bits.

This tree is what lets the decoder run in parallel. A pixel depends only on pixels of earlier steps, so the values of one step do not depend on each other: the GPU predicts them all at once, and interleaved streams of rANS~\cite{ans,interleaved}, a fast entropy coder, which we call lanes, decode their corrections in parallel, each lane carrying its state from one step to the next. Coarse-to-fine orders have long been used in image coding~\cite{hint,kimcho,kamisli}; what we add is an entropy coder designed together with the order, with lanes sized by the image, not by the step as in vectorised ANS~\cite{hilloc}, so that neither link of the chain grows with the image. Since the root's spacing is fixed in pixels, a photo of any size decodes in at most 72 steps, and no lane decodes more than a set number of values (\cref{tab:why}): the depth is fixed, though a larger photo has larger steps.

The tree also hosts classical and learned lossless coding in one format. It fixes which values are decoded together, but not how each is predicted or how its path is paid for, so hand-designed rules and neural networks become modes of the same format: a \emph{fast} mode stores the corrections of a reversible wavelet at fixed widths, with no entropy coder; a \emph{dense} mode predicts each pixel by the mean of four neighbours and codes the paths with histograms learned once, with no network; and \emph{neural} modes, with networks from 16 neurons in 2 layers (16\x2) to 768 in 4 (768\x4), give every possible correction its own probability, and the encoder tunes them to each photo. All decode bit-exactly on the GPU, and on other GPU types with our portable kernels.

Across this family, no codec we ran on the same GPU beats any \fmt mode on both size and speed, on Kodak or on two sets of large photos (\cref{fig:kodak,tab:family}). Without a network, \fmt beats NVIDIA's HTJ2K in both size and speed. With a small network, it makes smaller files than JPEG XL and decodes them faster on one GPU than JPEG XL does on a 32-thread CPU. With a large network, it decodes a Kodak photo 9 to 61\x\ faster than the learned codecs we ran from their authors' code on an A100, at the same size or smaller (\cref{tab:kodak}). At full load, the fast mode decodes more pixels per second than any other codec we ran (\cref{tab:throughput}).

\noindent In summary, we make three contributions:
\begin{enumerate}[leftmargin=*,itemsep=0pt,topsep=2pt]
\item \textbf{Where the time of learned lossless decoding goes.} Profiles of three released codecs, and our own networks decoded with longer coder chains or more steps, show that much of it goes to waiting, not computing (\cref{sec:ablations}, Sec.~S6.6).
\item \textbf{A decoding schedule whose sequential work does not grow with the image.} The order and the entropy coder are designed together: a coarse-to-fine tree of corrections, each a short path of decisions, fixes the number of steps by its root's spacing, and rANS lanes whose number grows with the image carry their states across steps, so prediction and entropy decoding both run in parallel on the GPU. Neither half suffices alone: with one coder, the same order still takes most of a second~\cite{kamisli}; with fresh lanes at every step, files grow by up to 14\% (about 24\% in the dense mode; \cref{sec:tree,sec:correct,sec:decoding,sec:why}, Sec.~S3.3).
\item \textbf{Bit-exact codecs, measured end to end.} GPU decoders for every mode, portable kernels whose streams decode alike on different GPUs (also with 8-bit tensor-core layers) and CPU decoders, compared from compressed bytes to verified pixels with classical codecs on one GPU and one CPU and with learned codecs run from their authors' code, including where they still win (\cref{sec:experiments,sec:limitations}).
\end{enumerate}

%% file: sec/related_work.tex
\section{Related work}
\label{sec:related}

\paragraph{Lossless coding as prediction.}
Most lossless codecs predict each pixel from decoded neighbours and code the residual: CALIC~\cite{calic} and LOCO-I, the core of JPEG-LS~\cite{locoi}, in raster order with fixed predictors and adaptive contexts; PNG~\cite{png}, WebP~\cite{webp} and QOI~\cite{qoi} with simpler rules; FLIF~\cite{flif} and JPEG XL~\cite{jpegxl,jxloverview} with a context tree learned for each image; GraLIC~\cite{gralic} mixes many predictors per pixel, far more slowly. JPEG 2000~\cite{jpeg2000} decorrelates the image with a reversible integer wavelet~\cite{lifting}, and HTJ2K~\cite{htj2k} codes it with a high-throughput block coder. Coarse-to-fine orders are old: hierarchical interpolation~\cite{hint,sp} predicts from coarser grids, Kim and Cho~\cite{kimcho} pair such a hierarchy with context-adaptive arithmetic coding, much as our dense mode does, and FLIF's interlaced mode, PNG's Adam7~\cite{png} and JPEG XL's Squeeze~\cite{jxloverview} decode coarse to fine.

\paragraph{Learned lossless codecs.}
PixelCNN~\cite{pixelcnn,pixelcnnpp} runs a network per value. Later codecs predict groups of values: multi- and subscale models~\cite{multiscale,spn,mspsm}, hierarchical latents (L3C~\cite{l3c}, SHVC~\cite{shvc}), super-resolution (SReC~\cite{srec}), frequency bands (LC-FDNet~\cite{lcfdnet}), lossy plus residual (RC~\cite{rc}, DLPR~\cite{dlpr}) and bit planes (ArIB-BPS~\cite{aribbps}). Integer flows~\cite{idf,ivpf,iflow,iodf}, probabilistic circuits~\cite{pcircuits}, per-image fitting (FNLIC~\cite{fnlic}, and for lossy coding Cool-chic~\cite{coolchic} and C3~\cite{c3}) and adaptation (\mbox{CALLIC}~\cite{callic}) differ; checkerboard contexts~\cite{checkerboard,elic} decode lossy latents in two steps.

\paragraph{Making learned codecs fast.}
The attempt most similar to \fmt is Kamisli's interpolation codec~\cite{kamisli}, which stores a sparse grid of pixels and predicts the finer grids from coarser ones with small networks: essentially our decoding order. Its network runs only about 15 times per image, yet it still takes most of a second per photo by its paper, because a single arithmetic coder on the CPU decodes every value in turn. HPAC~\cite{hpac} bounds its network steps by working on patches and comes closest to our largest networks in size, but by its paper it takes about a second per Kodak photo, 17 times as long as our 512\x4. PILC~\cite{pilc} takes the opposite route: it runs entirely on the GPU and is fast, but its files are about as large as WebP's, which our dense mode, without any network, beats in both size and throughput (\cref{tab:throughput}). Wavefront codecs~\cite{dlic,gumus} decode many pixels at once, DLIC~\cite{dlic} with rANS on the GPU, but in steps whose number grows with the image (\cref{sec:ablations}). Together they show that a parallel order alone does not make decoding fast, and that speed bought with a simple model gives up the size. For lossy video, DCVC-RT~\cite{dcvcrt} finds speed limited by operations other than computation and trims them in its network; \fmt instead bounds the number of sequential steps. What \fmt adds is an entropy coder designed together with its order: the values of each step are spread over interleaved rANS lanes that carry their states from step to step, and the predictors are cheap enough per step, with integer probabilities, to keep both prediction and entropy decoding on the GPU, bit-exact. Unlike JPEG XL and HTJ2K, which decode in parallel by coding blocks of the image apart, so that no value sees across a block's border, \fmt splits no image into blocks.

\paragraph{Parallel entropy decoding and exactness.}
Classical formats already decode on GPUs: NVIDIA's libraries decode JPEG 2000, HTJ2K~\cite{nvjpeg2000}, TIFF~\cite{nvimagecodec} and general-purpose formats~\cite{nvcomp}. Huffman~\cite{gpuhuff} and ANS~\cite{ans,gpuans,recoil,dietgpu} streams decode in parallel once split into parts or interleaved~\cite{interleaved}, and vectorised ANS brings this to learned codecs~\cite{hilloc}. These methods parallelise the coder once the probabilities of many values are known; in a learned codec, the decoding order decides how many are known at once, and the tree of \fmt makes them every value of a step. As in vectorised ANS, our lanes keep their states across stages, but their number follows the image, not the step, so steps of 96 to 98,304 values share them (\cref{sec:decoding}). Integer networks~\cite{intnets} make probabilities identical across machines, as do our integer probability tables and portable kernels, and instance-adaptive compression~\cite{rozendaal} counts a per-image update in the rate, as we do.

%% file: sec/method.tex
\section{FMT: an image as a tree of corrections}
\label{sec:method}

\fmt decodes an image the way one sketches a picture: the layout first, the details last.

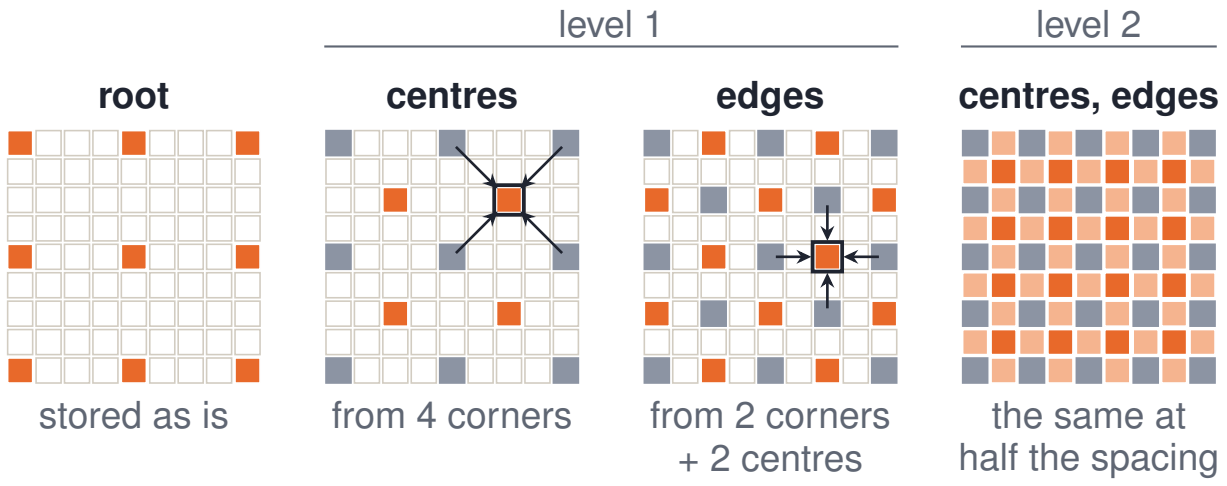
\begin{figure}[t]
\centering
\input{fig/tree_tikz}
\vspace{-2mm}
\caption{\textbf{How the tree fills an image}, on a $9\times9$ grid with a root every 4th pixel (\fmt: every 64th,
with 6 levels). Grey: decoded; orange: added in this step (level 2: centres dark, edges light).}
\label{fig:tree}
\vspace{-3mm}
\end{figure}
\subsection{A tree of pixels}
\label{sec:tree}

\fmt decodes an image from coarse to fine, as \cref{fig:tree} shows on a small grid. It first stores a sparse grid of pixels, the \emph{root}: one pixel every $S=64$ pixels in each direction, stored as is. It then adds the remaining pixels in $L=6$ levels, each of which halves the spacing of the decoded grid.
A level sees the decoded pixels as the corners of squares and fills each square in two steps. First it adds the square's \emph{centre}, which has decoded pixels at its four diagonal corners. Then it adds the midpoint of each side, an \emph{edge}, which now has decoded pixels on all four sides: the two corners of that side and the two centres next to it. The decoded pixels then form a grid twice as fine, which the next level fills in the same way, until every pixel is decoded. In coordinates, level $\ell$ fills squares of side $s=S/2^{\ell-1}$ and adds the pixels at half that spacing, $h=s/2$:
\begin{align}
\text{centres:}\;& (y,x)\equiv(h,h) \pmod s, \notag\\
\text{edges:}\;& (y,x)\equiv(0,h) \text{ or } (h,0) \pmod s. \label{eq:levels}
\end{align}
Each level adds three times as many pixels as were already decoded: in a $768\times512$ photo, the root holds 96 pixels, the first level adds 288 and the last 294,912. Two properties follow. Every new pixel has decoded neighbours on all sides, and the pixels added in one step never depend on each other, so a GPU can process them all at once (\cref{sec:decoding}). A neighbour that falls outside the image is reflected into it or replaced by the nearest decoded pixel, so the sides need not be multiples of 64. We call the structure a tree after the quadtree of its squares: every pixel is reached from the root through the levels above it, and its dependencies, on several coarser pixels, form a directed acyclic graph.

\subsection{Predict, then correct}
\label{sec:correct}

Every new value $v$ is first predicted, as $\hat v$, from pixels decoded in earlier steps, and the file stores only the correction (residual) $r=v-\hat v$ (Fig.~S2). The finer the level, the closer the neighbours and the smaller the corrections. A correction is binarised into a short path of discrete decisions, as in Exp-Golomb codes~\cite{expgolomb}, JPEG's magnitude categories~\cite{jpeg} and JPEG XL's hybrid integers~\cite{jxloverview}. The sign is folded in first, as in LOCO-I~\cite{locoi}: $u=2r$ if $r\ge0$ and $u=-2r-1$ otherwise, so small corrections of either sign become small numbers. The first decision is the size of $u$, its number of bits, cheap to code as small sizes dominate; the bits after its leading one follow. The modes differ in how they pay for these decisions (\cref{sec:modes}): the fast mode stores them at fixed widths, the dense mode codes them with shared histograms, and the neural modes give every possible correction its own probability. Before coding, all modes but the fast one convert the colours with one of JPEG XL's reversible transforms~\cite{jxloverview}, chosen by the sum of $\log_2(1+|e|)$ over the residuals $e$ of LOCO-I's predictor~\cite{locoi}.

\subsection{Three ways to code the tree}
\label{sec:modes}

\begin{table}[t]
\centering
\footnotesize
\setlength{\tabcolsep}{4.2pt}
\begin{tabular}{@{}llll@{}}
\toprule
& \textbf{Dense} & \textbf{Neural} & \textbf{Fast} \\
\midrule
Tree & pixels, 36 steps & pixels, 72 steps & 5/3 wavelet \\
Colours & ranked first & best of top 4 & YCoCg-R \\
Prediction & mean of 4 & mean of 4 & lifting \\
Probabilities & 256 histograms & 5 logistics & none \\
Model input & 2 context values & 24--48 neighbours & --- \\
Networks & --- & 16\x2 to 768\x4 & --- \\
Cost per value & integer ops & 2.6\,k--1.9\,M MACs & integer ops \\
Lane length $T$ & 512 or 2,048 & 2,048--8,192 & --- \\
Per photo & histogram choice & update, 0.14--1.2\% & --- \\
\bottomrule
\end{tabular}
\vspace{-2mm}
\caption{\textbf{The three modes of \fmt.} Steps: per photo, three channels. Colours: one of JPEG XL's 42 reversible transforms, by a fast estimate. Networks: neurons \x\ layers, one per channel; MACs: multiply--adds per coded value. $T$: values per lane (\cref{sec:decoding}). Per photo: side information besides the corrections (update: file share).}
\label{tab:modes}
\vspace{-3mm}
\end{table}

An entropy coder turns each value into bits according to the probability the decoder will give it: a value of probability $p$ costs about $-\log_2 p$ bits, so the better the probabilities, the smaller the file. The encoder and the decoder must therefore compute the same probabilities from what is already decoded. Classical codecs do this by counting: they sort values into a few \emph{contexts}, classes of similar neighbourhoods, and code each value with the histogram of its context~\cite{calic,locoi}. Learned codecs instead let a network compute the probability of every possible value from the neighbourhood~\cite{pixelcnnpp,l3c}. Both hand their probabilities to an entropy coder such as rANS~\cite{ans}, whose decoder keeps one integer state and reads each value by finding where that state falls in the value's cumulative distribution. \fmt offers both kinds, and a third that needs no probabilities at all (\cref{tab:modes}). The tree fixes which values are decoded together; the modes decide how each value is predicted and how its path is paid for.

\paragraph{Dense: shared histograms.}
The dense mode follows the classical recipe on the pixel tree. It predicts each new pixel by the rounded mean of its four nearest decoded neighbours (\cref{fig:tree}) and codes the first decisions of each correction, its size and the bit after its leading one, with rANS under a histogram; the remaining bits, close to uniform, are coded at uniform probability. The histogram depends on a context made of two values: how much the four neighbours differ, and how large their corrections were, there and in the previous colour channel. Similar neighbourhoods have similar histograms in every image, so one codebook of 256 histograms, learned once on training photos, serves all images, and a file stores only which histogram each of its contexts uses. Decoding a value takes a few integer operations and no network. Its limit is the context: two hand-chosen values can tell only so many neighbourhoods apart.

\paragraph{Neural: learned probabilities.}
The neural modes remove that limit by replacing the context and its histogram with a network that computes the probability of every possible correction from the neighbourhood itself. For each value, a small multilayer perceptron reads the nearest decoded pixels, relative to the mean of the nearest four, which remains the prediction; how much these four differ; and, in the second and third colour channels, the values already decoded at and around the same pixel. One-hot inputs tell it the level and the kind of step, so one network per colour channel serves the whole tree. It outputs a mixture of discretised logistics~\cite{pixelcnnpp}, a smooth distribution over all corrections, which the encoder and the decoder turn into the same integer table for rANS (\cref{sec:decoding}). The network only prices each correction: a weaker one makes a larger file, never a wrong pixel. The networks are trained to minimise the code length of the corrections on crops of training photos. For more context, the neural modes split each step in two, so the second half also sees the first: the centres in two checkerboard halves, the edges as horizontal, then vertical midpoints.

\paragraph{Tuning to each photo.}
A network trained on many photos is not the best one for a given photo. As JPEG XL learns a context tree for each image, the encoder of the neural modes adapts the network to the photo and stores the change in the file: a scale and a shift for each hidden neuron, the output biases, and a change of the output layer within a fixed random subspace (as in FiLM~\cite{film} and LoRA~\cite{lora}). A few hundred gradient steps minimise the photo's code length plus a penalty standing in for the change's coded size; the change is then rounded and entropy-coded, in a small share of the file (\cref{tab:modes}). The decoder merges it into the weights once, so a tuned photo decodes as fast as an untuned one. The encoder also tries the best-ranked colour transforms and the 8 orientations, and keeps the smallest file (Sec.~S4).

\paragraph{Fast: no entropy coder.}
When speed matters above all, the fast mode drops probabilities altogether: every correction is stored at a fixed width, so the decoder only reads bits. Instead of the pixel tree, it uses the tree of the reversible integer 5/3 wavelet of JPEG~2000~\cite{lifting,jpeg2000}, which, like the pixel tree, predicts new coefficients from coarser ones and keeps only the corrections. The corrections are stored in groups of $8\times8$. Each group first stores, in one byte, the largest size among its corrections and the range of their sizes; each correction then stores how far its size lies below the largest, at the width that range implies, and its remaining bits follow. Since every field has a width known from its group's byte, the decoder finds where each group and each correction begins with prefix sums, running totals that a GPU computes in parallel, and decodes all groups at once.

\subsection{Decoding in parallel, exactly}
\label{sec:decoding}

\begin{figure}[t]
\centering
\includegraphics[width=\linewidth]{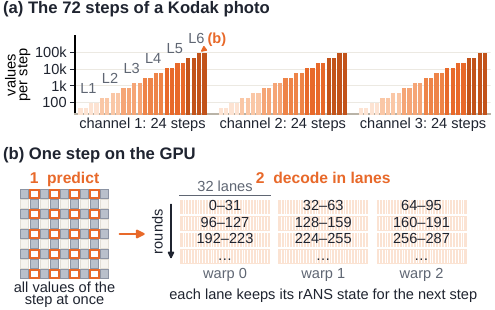}
\vspace{-6mm}
\caption{\textbf{Decoding in parallel.} (a) Values per step of a Kodak photo in the neural modes: per channel and level, two checkerboard halves of the centres, then horizontal and vertical edges. (b) One step: all its values are predicted at once; value $j$ then goes to warp $\lfloor j/32\rfloor \bmod G$, lane $j \bmod 32$ (here $G=3$), and each lane decodes its values round by round, keeping its rANS state for the next step. For the 512\x4 ($G=5$), the step marked in (a) takes 615 rounds, and no lane decodes more than 7,410 values in the photo.}
\label{fig:schedule}
\vspace{-3mm}
\end{figure}

\fmt decodes an image one step at a time, and everything inside a step in parallel (\cref{fig:schedule}).

\paragraph{One step, all its values at once.}
Within a step, no value depends on another (\cref{sec:tree}): a centre depends only on coarser pixels, an edge also on the centres of its level, the second half of a split step also on the first, and a later colour channel also on the earlier ones. The GPU therefore decodes a step in two phases: it predicts all its values and their probabilities at once, then decodes their corrections in parallel (\cref{fig:schedule}b). With the three colour channels decoded in turn, a photo of any size takes 36 steps in the dense mode and 72 in the neural modes (\cref{fig:schedule}a). This fixes the number of steps, not the time: a larger image has larger steps.

\paragraph{Lanes that carry their state.}
An rANS decoder is sequential: it reads one value after another from its stream. To decode many values at once, the stream is interleaved: the values are dealt to many \emph{lanes}, each a separate rANS decoder with its own state, run by one GPU thread~\cite{interleaved}. \fmt deals the values of every step to the lanes in turn, and each lane keeps its state from one step to the next. The number of lanes therefore follows the size of the image, not of the step: an image gets enough lanes that each decodes about $T$ values in all, a lane length that trades parallelism against the few bytes of each lane's final state (\cref{tab:modes}). Giving each step lanes of its own would need many times as many lanes for the same chain length, each with a final state to store, and make files 13\% larger with our 16\x2 (Sec.~S3.3).

\paragraph{Bit-exact probabilities.}
Lossless decoding needs the decoder to compute exactly the same probabilities as the encoder: a single difference in rounding makes it read a wrong value, and the error spreads to every value predicted from it. Floating-point results depend on the order of operations, which compilers, libraries and GPUs may change. We therefore turn each network's mixture into an integer cumulative distribution, with fixed-point arithmetic and a table of the sigmoid, in which every value keeps a nonzero frequency (Sec.~S5.1), and our kernels fix the order of every operation. The dense and fast modes compute with integers only, so any processor decodes their streams, and those of our small networks decode bit for bit on all three GPUs we tested.

\subsection{Why a tree?}
\label{sec:why}

Two things decide how fast an image decodes and how small its file is: how many steps must run one after another, and how much of the decoded image each value can see. The tree trades a little of the second for much of the first
(\cref{tab:why}).

\paragraph{Few steps.}
In raster order, every value waits for the one before it; in the tree, only the steps wait for each other, and the lanes keep each decoder's chain short. The lanes are not tied to the tree: any order that decodes groups of values could use them. What the tree adds is that these groups stay few and large for an image of any size, because its root is so sparse that storing it costs almost nothing.

\paragraph{Context on all sides.}
Most pixels see more in the tree than in raster order: the three quarters of the pixels that the last level adds have decoded neighbours on all four sides, where raster order offers only those above and to the left. The price is paid at the coarse levels, whose decoded neighbours are far away: their values cost about twice the bits, but the first three levels hold under 2\% of values. Against a wavefront, it costs small networks 0.3 to 2.4\% of size (\cref{sec:ablations}).

%% file: fig/tree_tikz.tex
\definecolor{treeOld}{HTML}{8C94A3}
\definecolor{treeNew}{HTML}{E8692A}
\definecolor{treeNewB}{HTML}{F5B48F}
\definecolor{treeEmpty}{HTML}{D2CCC1}
\definecolor{treeInk}{HTML}{1F2430}
\definecolor{treeMuted}{HTML}{5F6673}
\newcommand{\treegrid}[3]{%
  \foreach \y in {0,...,8} {\foreach \x in {0,...,8} {%
    \pgfmathtruncatemacro{\k}{(mod(\y,4)==0 && mod(\x,4)==0) ? 0 :
      ((mod(\y,2)==0 && mod(\x,2)==0) ? ((mod(\y,4)==2 && mod(\x,4)==2) ? 1 : 2) :
      ((mod(\y,2)==1 && mod(\x,2)==1) ? 3 : 4))}%
    \pgfmathtruncatemacro{\nxt}{#2+1}%
    \ifnum\k<#2 \def\fc{treeOld}\def\ec{treeOld}%
    \else\ifnum\k=#2 \def\fc{treeNew}\def\ec{white}%
    \else\ifnum\k=\nxt \ifnum#3=1 \def\fc{treeNewB}\def\ec{white}\else\def\fc{white}\def\ec{treeEmpty}\fi
    \else \def\fc{white}\def\ec{treeEmpty}\fi\fi\fi
    \filldraw[fill=\fc, draw=\ec, line width=0.3pt] (#1+\x+0.06, \y+0.06) rectangle (#1+\x+0.94, \y+0.94);
  }}%
}
\newcommand{\treearrow}[5]{%
  \draw[-{Stealth[length=2.8pt, width=2.6pt]}, line width=0.5pt, treeInk, shorten <=1pt, shorten >=3pt]
    (#1+#3+0.5, #2+0.5) -- (#1+#5+0.5, #4+0.5);}
\begin{tikzpicture}[x=0.188cm, y=-0.188cm, font=\sffamily\scriptsize]
  \def\ga{0}\def\gb{11.2}\def\gc{22.4}\def\gd{33.6}
  \treegrid{\ga}{0}{0}
  \treegrid{\gb}{1}{0}
  \draw[treeInk, line width=0.7pt] (\gb+6-0.02, 2-0.02) rectangle (\gb+7.02, 3.02);
  \foreach \sy/\sx in {0/4, 0/8, 4/4, 4/8} {\treearrow{\gb}{\sy}{\sx}{2}{6}}
  \treegrid{\gc}{2}{0}
  \draw[treeInk, line width=0.7pt] (\gc+6-0.02, 4-0.02) rectangle (\gc+7.02, 5.02);
  \foreach \sy/\sx in {4/4, 4/8, 2/6, 6/6} {\treearrow{\gc}{\sy}{\sx}{4}{6}}
  \treegrid{\gd}{3}{1}
  \foreach \o/\t/\n in {\ga/root/{stored as is}, \gb/centres/{from 4 corners},
                        \gc/edges/{from 2 corners\\+ 2 centres}, \gd/{centres, edges}/{the same at\\half the spacing}} {
    \node[anchor=base, font=\sffamily\scriptsize\bfseries, text=treeInk, inner sep=1pt] at (\o+4.5, -0.75) {\t};
    \node[anchor=north, align=center, text=treeMuted, inner sep=1pt] at (\o+4.5, 9.35) {\n};
  }
  \draw[treeMuted, line width=0.4pt] (\gb, -2.9) -- (\gc+9, -2.9)
    node[midway, anchor=south, inner sep=1.5pt, text=treeMuted] {level 1};
  \draw[treeMuted, line width=0.4pt] (\gd, -2.9) -- (\gd+9, -2.9)
    node[midway, anchor=south, inner sep=1.5pt, text=treeMuted] {level 2};
\end{tikzpicture}

%% file: sec/experiments.tex
\section{Experiments}
\label{sec:experiments}

\begin{figure*}[t]
\centering
\includegraphics[width=\linewidth]{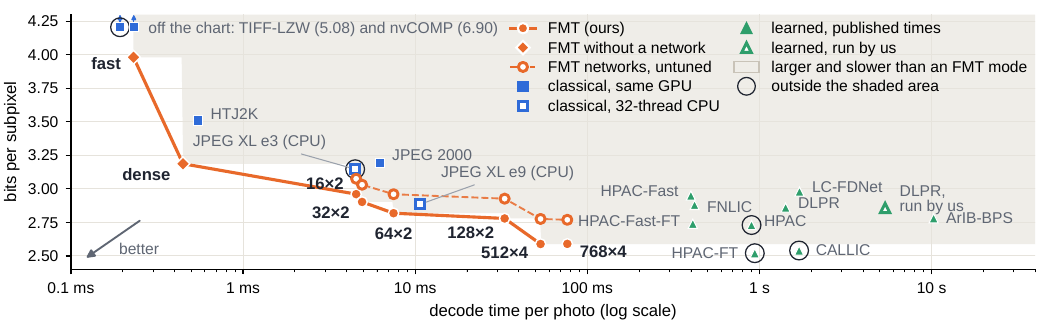}
\vspace{-6mm}
\caption{\textbf{Kodak photos}: size against decode time for \fmt (diamonds: no network; circles: networks tuned to each photo; hollow circles: untuned, at the times of the tuned streams) and the classical codecs on one RTX 4090 (hollow squares: JPEG XL on a 32-thread CPU), and for learned codecs at their published times~\cite{hpac,callic} (our A100 runs of FNLIC, LC-FDNet and ArIB-BPS: \cref{tab:kodak}). Every codec in the shaded area is larger \emph{and} slower than an \fmt mode, or as fast (nvCOMP); the ringed ones are not: TIFF-LZW (faster, 28\% larger), HPAC-FT and \mbox{CALLIC} (smaller, at 18 and 32\x\ the time of our 512\x4), JPEG XL e3, and HPAC (1.7\% smaller than our untuned 512\x4).}
\label{fig:kodak}
\vspace{-2mm}
\end{figure*}

\begin{table}[t]
\centering
\footnotesize
\setlength{\tabcolsep}{3.2pt}
\begin{tabular}{@{}lrrrrrr@{}}
\toprule
& \multicolumn{2}{c}{Kodak} & \multicolumn{2}{c}{CLIC.p / 4} & \multicolumn{2}{c}{DIV2K / 4} \\
\cmidrule(lr){2-3}\cmidrule(lr){4-5}\cmidrule(l){6-7}
Codec & bpsp & ms & bpsp & ms & bpsp & ms \\
\midrule
\multicolumn{7}{@{}l}{\emph{\fmt on one RTX 4090}} \\
fast$^\star$ & 3.981 & 0.23 & 3.870 & 0.71 & 4.145 & 0.87 \\
dense$^\star$ & 3.187 & 0.45 & 2.671 & 0.83 & 3.071 & 1.02 \\
16\x2$^\star$ & 2.960 & 4.5 & 2.447 & 7.9 & 2.782 & 8.9 \\
32\x2$^\star$ & 2.902 & 4.9 & 2.408 & 8.9 & 2.731 & 10.0 \\
64\x2$^\star$ & 2.818 & 7.5 & 2.364 & 16.9 & 2.679 & 19.1 \\
128\x2 & 2.779 & 33.1 & -- & -- & -- & -- \\
512\x4 & \best{2.589} & 53.5 & 2.239 & 233 & 2.545 & 270 \\
768\x4 & \best{2.589} & 76.5 & \best{2.236} & 372 & 2.540 & 432 \\
\midrule
\multicolumn{7}{@{}l}{\emph{Others on the same GPU, and JPEG XL on the CPU}} \\
nvCOMP Bitcomp & 6.904 & 0.23 & 6.352 & \best{0.45} & 6.696 & \best{0.50} \\
TIFF-LZW (nvTIFF) & 5.078 & \best{0.19} & 4.226 & 0.52 & 5.204 & 0.77 \\
HTJ2K & 3.510 & 0.55 & 2.960 & 1.30 & 3.354 & 1.54 \\
JPEG 2000 & 3.191 & 6.2 & 2.760 & 7.5 & 3.132 & 9.0 \\
DLPR, run by us & 2.857 & 5373 & 2.241 & 14280 & \best{2.522} & 16189 \\
JPEG XL e9 (CPU) & 2.885 & 10.7 & 2.457 & 25.1 & 2.785 & 29.8 \\
\bottomrule
\end{tabular}
\vspace{-2mm}
\caption{\textbf{The \fmt family}: bits per subpixel and decode time per image, one image at a time, on Kodak and every fourth image of CLIC.p and DIV2K; networks tuned to each photo, update included (DLPR: one model for all images). Bold: the smallest and the fastest per column pair. Lanes of 512 values for dense, 2,048 for 16\x2 to 128\x2, 8,192 for 512\x4 and 768\x4; the 128\x2 was not timed on the large photos. $^\star$: streams decoded exactly, with these kernels, on the two other GPU setups we tried (\cref{sec:decoding}).}
\label{tab:family}
\vspace{-3mm}
\end{table}

\subsection{Setup}
\label{sec:setup}

\paragraph{Data.}
We time every codec on the 24 Kodak photos~\cite{kodak} ($768\times512$) and on every fourth of the 41 CLIC 2020 professional (CLIC.p)~\cite{clic} and 100 DIV2K~\cite{div2k} validation photos (2.4 and 2.8 megapixels on average), so that even the slowest codecs decode the same images; \cref{tab:fullsets} gives sizes on the full sets. The CLIC.p subset is easier than its full set for every codec; the DIV2K subset is within 0.5\% of its own. Our networks are trained on crops of the DIV2K, CLIC and Flickr2K~\cite{flickr2k} training images (Sec.~S3.1), so no test image was seen in training. Every size is that of a complete file.

\paragraph{Codecs and machines.}
The GPU codecs run on one NVIDIA RTX 4090, whose host is an AMD EPYC 7702. There we compare with HTJ2K and JPEG~2000 through nvJPEG2000~\cite{nvjpeg2000}, TIFF-LZW through nvTIFF~\cite{nvimagecodec}, and eight nvCOMP~\cite{nvcomp} compressors on raw and PNG-filtered pixels. The CPU codecs run on an Intel Core i9-14900K with 32 threads: PNG, QOI, WebP, JPEG-LS, JPEG~2000, AVIF, and JPEG XL at five efforts from 1 to 9 (and 10 on Kodak), each at its best thread count. For learned codecs we use the decode times that HPAC's authors measured on an RTX 4090D~\cite{hpac} and \mbox{CALLIC}'s own~\cite{callic}. We also ran four learned codecs with their authors' code and models: DLPR~\cite{dlpr} on our RTX 4090, and FNLIC, LC-FDNet and ArIB-BPS on an A100 (Secs.~S6.2 and~S6.6).

\paragraph{Implementation.}
In each step, the 16\x2 to 64\x2 networks run whole in one fp32 kernel, one value per thread; the larger ones run a feature kernel and bfloat16 matrix multiplications, with an fp32 output layer. On the CPU, \fmt decodes the fast mode with OpenMP and 8-bit 32\x2 and 64\x2 networks with AVX-VNNI, from their own streams.

\paragraph{What we time.}
A decode starts from the compressed bytes in host memory and ends with the RGB image in GPU memory (CPU codecs: in host memory), with parsing, the merge of the per-photo update and the inverse colour transform inside the clock. We report the minimum over 7 rounds after a warm-up, averaged over the images, and check every decode: every image we timed decoded bit for bit. Medians change no comparison (Tab.~S22), and Tab.~S14 gives the protocol of every result. Index tables that depend only on the image size, such as each value's neighbours, are built once per size, outside the clock; they add about a millisecond to the first decode of a Kodak-sized photo, and at most 1.4\,ms per photo on 5,000 COCO photos of 603 sizes (Secs.~S7.2 and~S7.3). We will release the code, models, streams and logs, from which one script recomputes every reported number.

\subsection{One family along the frontier}
\label{sec:results}

\cref{fig:kodak} and \cref{tab:family} show the main result: on the GPU, no codec beats any \fmt mode on both size and speed, on any of the three datasets (large photos: Fig.~S1). Only general-purpose formats with much larger files decode faster than our fast mode, and only learned codecs that take tens of times longer make smaller files: HPAC-FT and \mbox{CALLIC} on Kodak, and DLPR and ArIB-BPS on DIV2K, by 0.7 and 1.0\%.

\begin{table}[t]
\centering
\footnotesize
\setlength{\tabcolsep}{3.4pt}
\begin{tabular}{@{}lrrlrr@{}}
\toprule
Learned codec & bpsp & time & \fmt & size & faster \\
\midrule
\multicolumn{6}{@{}l}{\emph{One model for all images}} \\
HPAC-Fast & 2.95 & 0.40\,s & 512\x4$^\circ$ & $-$5.9\% & 7.5\x\ \\
HPAC & 2.73 & 0.90\,s & 512\x4$^\circ$ & $+$1.7\% & 17\x\ \\
DLPR & 2.86 & 1.42\,s & 512\x4$^\circ$ & $-$3.0\% & 27\x\ \\
LC-FDNet$^\ddagger$ & 2.94 & 1.00\,s & 512\x4$^\circ$ & $-$5.6\% & 17\x\ \\
ArIB-BPS$^\ddagger$ & 2.78 & 3.6\,s & 512\x4$^\circ$ & $-$0.2\% & 61\x\ \\
\multicolumn{6}{@{}l}{\emph{Adapted to each image}} \\
FNLIC$^\ddagger$ & 2.88 & 0.52\,s & 512\x4 & $-$10.2\% & 8.8\x\ \\
HPAC-Fast-FT & 2.74 & 0.41\,s & 512\x4 & $-$5.5\% & 7.7\x\ \\
HPAC-FT & 2.52 & 0.94\,s & 512\x4 & $+$2.7\% & 18\x\ \\
\mbox{CALLIC} & 2.54 & 1.7\,s & 512\x4 & $+$1.9\% & 32\x\ \\
\bottomrule
\end{tabular}
\vspace{-2mm}
\caption{\textbf{Head to head with learned codecs on Kodak}, at their published times (RTX 4090D; \mbox{CALLIC}: GPU not stated): the size of our 512\x4 relative to each (negative: ours is smaller) and its speed-up (ours: 53.5\,ms on one RTX 4090), untuned ($^\circ$: 2.7755 \bpsp, the tuned stream's time, an upper bound: \cref{sec:modes}) against codecs with one model for all images, tuned against the others. -Fast, -FT: HPAC's faster and per-image variants. $^\ddagger$: run by us with their authors' code on one A100, against our 512\x4 there (58.7\,ms); FNLIC without its 17\,ms of deleting files.}
\label{tab:kodak}
\vspace{-3mm}
\end{table}

\paragraph{Without a network.}
The dense mode has no network, only a fixed rule and 256 histograms learned once, yet it beats HTJ2K, decoded by NVIDIA's library, on both axes and on every dataset: its files are 8--10\% smaller, it decodes a photo 1.2--1.6\x\ faster (1.1--1.4\x\ with a faster host CPU; Sec.~S7.4), and, with longer lanes, it stays smaller and faster at full load (\cref{tab:throughput}). It is 9--14\x\ faster than JPEG~2000 on the same GPU, at the same size or smaller.
\paragraph{Against JPEG XL.}
At effort 9, the highest we ran on all three datasets, JPEG XL (libjxl 0.11.2) is the strongest classical codec in our comparison. Tuned to each photo, as JPEG XL learns a context tree for each, our 64\x2 network makes files 2.3--3.8\% smaller, smaller on 45 of the 49 Kodak and DIV2K/4 photos (Tab.~S16), and decodes them 1.4--1.6\x\ faster (also at effort 10: Tab.~S22); at full load, our 16\x2 matches its size on the large photos at about 3.5\x\ its throughput (\cref{tab:throughput}). JPEG XL has no GPU decoder we know of, so these comparisons put one GPU against one CPU, each fully used. With libjxl 0.12 and 0.13 too, no JPEG XL setting, faster decoding included, is smaller and faster than any \fmt mode on any dataset (Sec.~S6.10).

\paragraph{Against learned codecs.}
\cref{tab:kodak} sets our 512\x4 against every learned codec with a published Kodak time. Matched on tuning, it is smaller than all of them but three, HPAC, HPAC-FT and \mbox{CALLIC}: it is 1.7--2.7\% larger than these, which take 17--32\x\ as long by their published times. Published times need not include entropy decoding and transfers, as ours do, and DLPR took 3.8\x\ its published time in our run, so we also ran FNLIC, LC-FDNet and ArIB-BPS beside ours on one A100. There, our 512\x4 decodes 8.8 to 61\x\ faster, and our tuned 64\x2, smaller than FNLIC, which also adapts to each photo, 41\x\ faster (Tab.~S17). FNLIC's released bitstreams decode wrongly on our GPU, all 24, while streams of our portable kernels decode alike on every GPU we tried and still decode 3.1\x\ faster than FNLIC (Sec.~S5.3).

\begin{table}[t]
\centering
\footnotesize
\setlength{\tabcolsep}{3.6pt}
\begin{tabular}{@{}lcc@{\hspace{12pt}}lcc@{}}
\toprule
\multicolumn{3}{@{}l}{\emph{Measured by us}} & \multicolumn{3}{l}{\emph{Published}~\cite{hpac,callic,llmvp}} \\
& CLIC.p & DIV2K & & CLIC.p & DIV2K \\
\midrule
\fmt 16\x2 & 2.602 & 2.775 & ArIB-BPS & 2.42 & 2.55 \\
\fmt 64\x2 & 2.506 & 2.673 & DLPR & 2.38 & 2.55 \\
\fmt 128\x2 & 2.480 & 2.644 & HPAC & 2.38 & 2.52 \\
\fmt 512\x4 & 2.386 & 2.551 & HPAC-FT$^*$ & 2.34 & 2.47 \\
\fmt 768\x4 & 2.382 & 2.546 & \mbox{CALLIC}$^*$ & 2.30 & 2.46 \\
JPEG XL e9 & 2.626 & 2.787 & LLM+VP & 2.25 & 2.29 \\
HTJ2K & 3.145 & 3.338 & & & \\
JPEG 2000 & 2.932 & 3.118 & & & \\
\bottomrule
\end{tabular}
\vspace{-2mm}
\caption{\textbf{Full validation sets} (41 CLIC.p, 100 DIV2K photos), bits per subpixel, our networks tuned to each photo. Ours: all bits over all subpixels of a set, as in the published tables, whose JPEG XL and JPEG 2000 rows on CLIC.p (2.63, 2.93) match ours, not the mean over photos (Sec.~S6.3). $^*$: adapted to each image, like ours; LLM+VP: LLM-based.}
\label{tab:fullsets}
\vspace{-3mm}
\end{table}

\paragraph{Full sets.}
On the full sets (\cref{tab:fullsets}), our tuned 768\x4 ties with the published sizes of DLPR and HPAC on CLIC.p and of DLPR on DIV2K, where HPAC is 1.0\% smaller; untuned, by what tuning saves on the subsets, it would trail them by about 1\% more (Tab.~S5). The codecs that adapt to each image remain 1.8--3.4\% smaller, and an LLM-based predictor~\cite{llmvp} 5.5--10\%, at 140--500\,s per image.

\subsection{Throughput}
\label{sec:throughput}

\begin{table}[t]
\centering
\footnotesize
\setlength{\tabcolsep}{2.2pt}
\begin{tabular}{@{}lrrr@{\hspace{5pt}}lrrr@{}}
\toprule
\multicolumn{4}{@{}l}{\emph{One RTX 4090}} & \multicolumn{4}{l}{\emph{i9-14900K, 32 threads}} \\
& Kodak & CLIC.p & DIV2K & & Kodak & CLIC.p & DIV2K \\
\midrule
\fmt fast & \best{8811} & \best{11792} & \best{9116} & QOI & \best{3077} & \best{3195} & \best{3512} \\
\fmt dense & 6716 & 7886 & 8019 & WebP & 1664 & 1771 & 1690 \\
HTJ2K & 4533 & 7097 & 7013 & PNG & 1572 & 1813 & 1596 \\
nvTIFF & 4558 & 6105 & 5587 & JXL e1 & 886 & 977 & 977 \\
nvCOMP & 2205 & 3364 & 3139 & \fmt fast & 838 & 748 & 762 \\
\fmt 16\x2 & 397 & 490 & 490 & JPEG-LS & 687 & 782 & 700 \\
\fmt 64\x2 & 161 & 185 & 188 & JXL e9 & 135 & 142 & 135 \\
\bottomrule
\end{tabular}
\vspace{-2mm}
\caption{\textbf{Throughput at full load}, in MP/s, each codec at its best threads and
batch size. Files of \cref{tab:family}, except the dense mode (lanes of 2,048: 3.150 / 2.635 / 3.034
\bpsp) and nvCOMP on CLIC.p (GDeflate). JXL: JPEG XL. Every codec: Tab.~S24.}
\label{tab:throughput}
\vspace{-3mm}
\end{table}

A training job decodes many images at once. At full load (\cref{tab:throughput}), the fast mode decodes 8.8 to 11.8 gigapixels per second, more than any codec we ran (Sec.~S9). Large networks serve latency, not bulk decoding: one photo fills the GPU for our 512\x4 (7\,MP/s, Tab.~S24). On a CPU, one image at a time, our fast mode is fastest, and our 8-bit 64\x2 is 2.1\% smaller than JPEG XL e9 at 1.6\x\ its time (Fig.~S1c).

\subsection{Design choices}
\label{sec:ablations}

\begin{figure}[t]
\centering
\includegraphics[width=\linewidth]{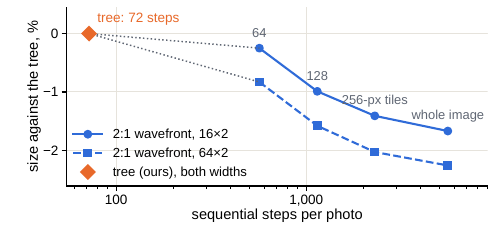}
\vspace{-6mm}
\caption{\textbf{Size against sequential steps}, Kodak: the 2:1 wavefront, whole-image and tiled, against the tree
with networks of the same width (tuned ideal code length, one session; Sec.~S3.4).}
\label{fig:steps}
\vspace{-3mm}
\end{figure}

\paragraph{Decoding order.}
To see what the tree costs, we train the same networks for a 2:1 wavefront, the most parallel order in which a value sees its left, upper and upper-right neighbours (Sec.~S3.4). Our 16\x2 and 64\x2 then gain 1.6\% on Kodak (files) and 2.4\% (tuned code length), but the gain does not grow with capacity: a 512\x4 trained both ways with a short recipe gains nothing once tuned. The wavefront, however, takes 78 times as many steps, a number that grows with the image: about 24,000 for a 4K frame, where the tree still takes 72 (our 16\x2 decodes one in 33\,ms, Sec.~S3.4). Our kernels, built for a few large steps, decode the wavefront 40\x\ slower; a decoder built for small steps would narrow this gap, but synchronising its steps alone would take as long as our 16\x2's whole decode. Tiles stop the steps from growing, but not the cost of having many (\cref{fig:steps}, Sec.~S3.4). Inside tiles of 64 pixels, the most parallel choice, the wavefront still takes 567 steps, 8\x\ the tree's: with the same encoder and kernels, its 16\x2 makes files 0.3\% smaller than ours at 4.6\x\ the decode time (its 64\x2: 0.8\%, in tuned code length), and our 64\x2 is 4.5\% smaller in less than half that time. The tree thus trades 0.3 to 2.4\% of size for 8 to 78\x\ fewer steps.

\paragraph{Root, lanes and tuning.}
A root every 32 or 128 pixels instead of 64 changes the code length by at most 0.5\% (Sec.~S3.5), and a single warp of lanes instead of 18 makes files only 0.4\% smaller and every network over 20\,ms slower (Sec.~S3.2), 0.65\,$\mu$s for each value a lane decodes in turn: the second link of the chain, measured with the networks unchanged. Per-photo tuning saves 3.7--6.7\% on Kodak, against 0.3--0.7\% for the colour transforms and orientations, and still pays on $64\times64$ crops (Secs.~S4.1, S4.5).

%% file: sec/conclusion.tex
\section{Limitations}
\label{sec:limitations}

\fmt is built for 8-bit photos written once and read many times. Tuning to each photo takes seconds on the GPU, 3.5--8.9\,s per Kodak photo with our 512\x4 on an A100 (\mbox{HPAC-FT}: 8.45\,s~\cite{hpac}) and 2--14\,s per 2K photo with our fastest encoders (Sec.~S4.4), and on a CPU, \fmt is competitive only one image at a time. Our decoder keeps tables and buffers for each image size, 276--403 bytes per pixel, up to 9.6\,GB for a 33-megapixel image (Sec.~S7.1), or none if rebuilt per image (+1.5\,ms, Sec.~S7.3), and a $64\times64$ thumbnail still takes the 72 steps: 3.2\,ms with our 16\x2, against about 0.5\,ms for CPU codecs (Sec.~S4.5). The best codecs that adapt to each image remain 1.8--3.4\% smaller on the full sets. Our 128\x2 and larger networks have no CPU decoder yet, and their portable kernels cost time: 3.0--4.4\x\ in fp32, or up to 1.2\x\ with 8-bit layers (RTX 3090, 4090), for files at most 0.25\% larger (Secs.~S5.3--S5.4). Images of 10 to 16 bits (film masters, medical scans) need networks trained on them and wider frequency tables; we tested photographs only. Files need checksums: errors spread to finer levels.

\section{Conclusion}
\label{sec:conclusion}

Our results suggest that much of the time learned lossless codecs take can go to sequential work rather than to their networks. A coarse-to-fine schedule designed with its entropy coder bounds this work, to a few dozen steps and a set number of values per lane, for an image of any size. On such trees, from a quarter of a millisecond to tens of milliseconds per photo, a codec without a network beats HTJ2K on both axes, small networks reach the file sizes of some learned codecs at classical speed, and our largest networks spend their decode mostly on arithmetic, not on entropy decoding.

%% file: strefs.tex
\nocite{ans,aribbps,calic,callic,checkerboard,clic,cublas,dcvcrt,dietgpu,div2k,dlic,dlpr,elic,expgolomb,ffv1,flickr2k,flif,fnlic,gpuans,gpuhuff,gralic,gumus,hilloc,hint,hpac,htj2k,idf,iflow,igpt,interleaved,intnets,iodf,ivpf,jpeg,jpeg2000,jpegxl,jxloverview,kamisli,kimcho,kodak,l3c,lcfdnet,lifting,llmvp,locoi,mspsm,multiscale,nvcomp,nvimagecodec,nvjpeg2000,pcircuits,pilc,pixelcnn,pixelcnnpp,png,qoi,rc,recoil,rozendaal,shvc,sp,spn,srec,statetrees,webp}

%% file: supp/supp.tex
\clearpage
\setcounter{section}{0}\setcounter{figure}{0}\setcounter{table}{0}\setcounter{algorithm}{0}
\renewcommand{\thesection}{S\arabic{section}}\renewcommand{\theHsection}{S\arabic{section}}
\renewcommand{\thefigure}{S\arabic{figure}}\renewcommand{\theHfigure}{S\arabic{figure}}
\renewcommand{\thetable}{S\arabic{table}}\renewcommand{\theHtable}{S\arabic{table}}
\renewcommand{\thealgorithm}{S\arabic{algorithm}}\def\theHalgorithm{S\arabic{algorithm}}
\makeatletter\let\Hy@backout\@gobble\makeatother
\maketitlesupplementary

\section{Frontiers on large photos and on a CPU}
\label{S:sec:frontiers}

\begin{figure*}[h]
\centering
\includegraphics[width=\linewidth]{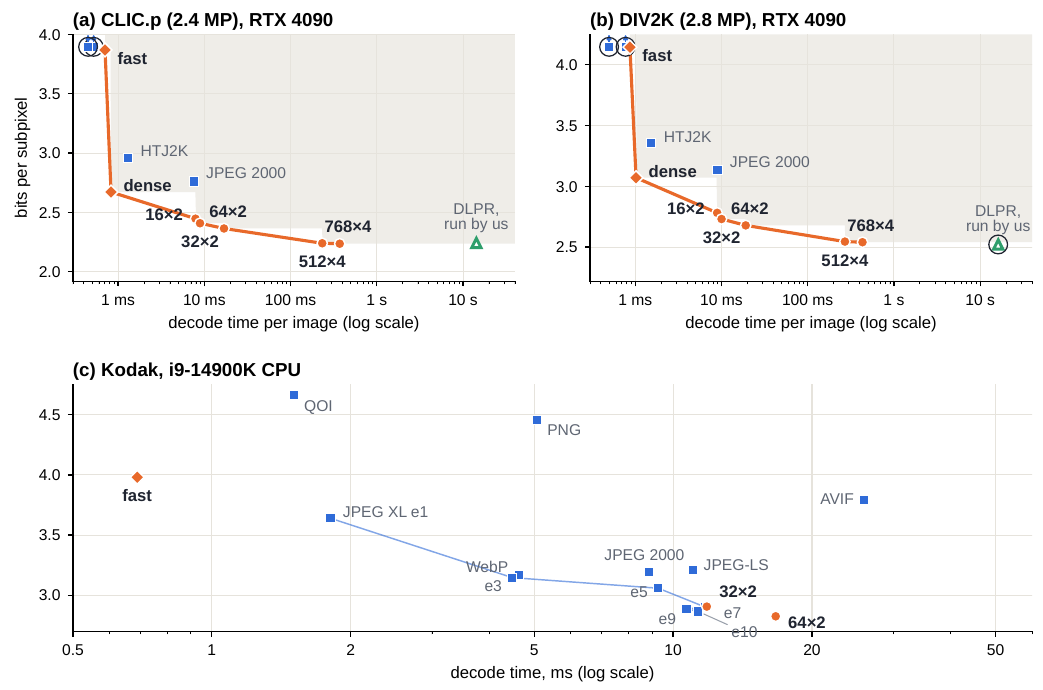}
\vspace{-6mm}
\caption{\textbf{Large photos, and a CPU.} (a, b) Every fourth CLIC.p and DIV2K validation photo on one RTX 4090; filled squares: classical codecs on the same GPU; hollow triangle: DLPR, run by us; the squares with arrows at the top edge are TIFF-LZW and nvCOMP, faster than our fast mode with larger files. In the shaded area every codec is larger and slower than an \fmt mode; the ringed ones are not. Of the codecs we ran on these subsets, only DLPR is below our curve, on DIV2K, by 0.7\% at 37\x\ our decode time; on the full sets, a few published learned codecs are smaller (main paper, Tab.~5). (c) Kodak on one i9-14900K CPU, every codec at its best number of threads: our fast mode (24 threads) opens a photo fastest, and our 8-bit 64\x2 is 2.1\% smaller than JPEG XL e9 at 1.6\x\ its decode time; in between, JPEG XL holds the frontier. JPEG-LS is shown with a re-measurement (11.06\,ms), JPEG XL e10 from a later session (\cref{S:tab:lat_all}).}
\label{S:fig:large}
\end{figure*}

Figure~\ref{S:fig:large} shows the size--speed frontiers of every codec on the two sets of large photos, one image at a
time on one RTX 4090, and on Kodak on one CPU; the main paper gives the numbers (Tab.~3) and the full-set sizes (Tab.~5).

\section{Decoding, step by step}
\label{S:sec:steps}

\begin{figure}[t]
\centering
\includegraphics[width=\linewidth]{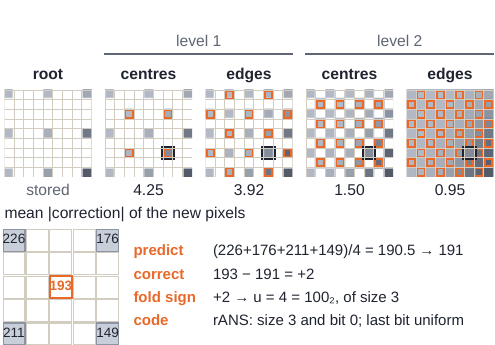}
\vspace{-6mm}
\caption{\textbf{A running example} (main paper, Fig.~2): a $9\times9$ crop (brightness), decoded with the dense-mode rule; for readability its root is every fourth pixel, so two levels complete it. Grey pixels are decoded, outlined ones are added by the step; the corrections shrink as the decoded neighbours get closer. Bottom: the marked pixel, predicted from its four decoded corners, and how its correction is coded.}
\label{S:fig:method}
\vspace{-3mm}
\end{figure}

\cref{S:fig:method} follows a $9\times9$ crop through two levels of the tree with the dense-mode rule, and one pixel
through its prediction and the coding of its correction. The main paper's Tab.~2 sets the three modes side by side. \cref{S:alg:decode} summarises the GPU decoder of the pixel-tree modes (main paper, Sec.~3.4). Phase 1 runs one thread per value of a step; phase 2 runs one thread per lane, lane $l$ of warp $g$, and the $G$ warps of the image keep their lanes' rANS states from step to step. The slot is the low 16 bits of a lane's state. In the dense mode, phase 1 computes the prediction and the context of each value, and phase 2 looks up its histogram and decodes the remaining bits of large corrections at uniform probability, through the same state.

\begin{algorithm}[h]
\caption{Decoding an image in a pixel-tree mode}
\label{S:alg:decode}
\footnotesize
\begin{algorithmic}[1]
\State read the header; merge the per-photo update into the weights
\State decode the root grid; load the states of the $32G$ lanes
\For{channel $c=1,2,3$}
  \For{step $k=1,\dots,K$} \Comment{$K=24$ (neural) or 12 (dense)}
    \State $n \gets$ number of values of step $k$
    \ForAll{$j<n$ \textbf{in parallel}} \Comment{predict}
      \State predict value $j$ from decoded neighbours
      \State 16 points of its cumulative distribution
    \EndFor
    \ForAll{lanes $(g,l)$ \textbf{in parallel}} \Comment{decode}
      \For{$j=32g+l$; $j<n$; $j \gets j+32G$}
        \State find the symbol that holds the state's slot
        \State update the state; the warp reads new words
        \State value $j \gets$ prediction $+$ correction
      \EndFor
    \EndFor
  \EndFor
\EndFor
\State undo the colour transform and the orientation
\end{algorithmic}
\end{algorithm}

\section{Design choices}
\label{S:sec:design}

\subsection{Training}
\label{S:sec:training}
The 16\x2 to 64\x2 train for 60,000 steps at a peak learning rate of $2\cdot10^{-3}$, then for 120,000 more at $10^{-3}$; the 128\x2 only for the first 60,000, which is why we did not time it on the large photos. The 512\x4 trains in three cycles, the last of 80,000 steps at a peak learning rate of $5\cdot10^{-4}$; the 768\x4 is the 512\x4 after its second cycle, widened without changing its function, then trained for the same third cycle. All train on the same 25,834 crops of $512\times512$ pixels: a grid over the DIV2K~\cite{div2k} training images and random crops of the CLIC~\cite{clic} and Flickr2K~\cite{flickr2k} training images, none of them a test image. Per-photo tuning runs Adam at a learning rate of $10^{-3}$ on a cosine schedule. The lane runs of this section and the encoder runs of \cref{S:sec:encoding} encode and decode the 24 Kodak photos on one RTX 3090 (CUDA 13.0, PyTorch 2.11), the GPU of our CPU machine, one image at a time, and every one of their decodes equals the photo bit for bit. Their times are those of the decoder's GPU part, from the parsed stream with the per-photo update merged to the decoded values before the inverse colour transform; the end-to-end times of the main paper include these steps.

\subsection{Lanes}
\label{S:sec:lanes}
\begin{figure*}[t]
\centering
\includegraphics[width=\linewidth]{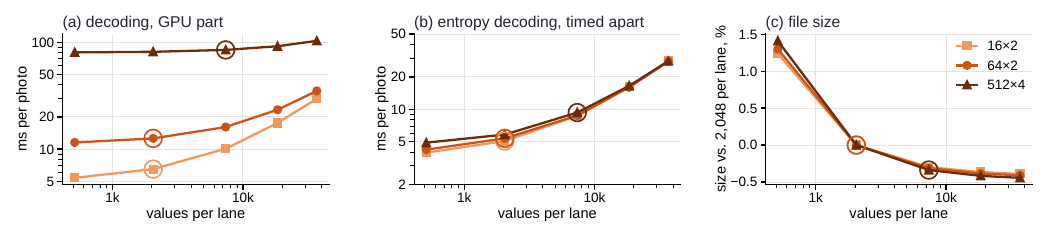}
\vspace{-6mm}
\caption{\textbf{What the lanes give}, for the same networks and coded values (\cref{S:tab:lanes}): (a) the GPU part of the decode of a Kodak photo, (b) its entropy decoding, timed apart, and (c) the size of the file, against the mean number of values per lane, on one RTX 3090. Rings: the lane lengths of our streams.}
\label{S:fig:lanes}
\end{figure*}

We pack the same coded values into lanes of about $T$ values each, from 512 to 262,144, which gives from $G=72$ warps down to a single one, and decode each stream in 7 rounds after a warm-up, interleaved over the lane lengths (\cref{S:tab:lanes}, \cref{S:fig:lanes}). Every value a lane decodes in turn costs about 0.65\,$\mu$s: from 2,048 values per lane to a single warp, the GPU part of the decode grows by 21.6 to 23.1\,ms with each network, and our 16\x2 then decodes a photo 4.6\x\ slower. A second decode, which waits after each kernel, times the network and the entropy decoding apart; the waiting adds 3--9\,ms, so its two parts sum to more than the decode, but its entropy part grows by the same 22--23\,ms while its network part stays within 0.3\,ms. The files shrink with longer lanes only because fewer lanes leave fewer final states to store: by 0.4\% from 2,048 values per lane to a single warp, while lanes of 512 values make them 1.2--1.4\% larger than lanes of 2,048.

\subsection{States carried across steps}
\label{S:sec:carry}
Our lanes keep their states from one step to the next. Without that, each step would need lanes of its own, each ending with its own final state. For the same chain, $T$ values per lane summed over the steps, steps of $n_s$ values then need at least $(\sum_s\sqrt{n_s})^2/T$ lanes in all, each step's in proportion to $\sqrt{n_s}$, against $N/T$ with carried states: 33.9 times as many for the 72 steps of the neural modes, at every image size, and 16.9 times for the 36 steps of the dense mode. We measured what so many lanes cost by packing the same untuned Kodak streams into 34 times as many carried lanes (lanes of 60 values instead of 2,048; for the 512\x4, 242 instead of 8,192, 31 times as many), every decode checked: each extra lane costs 25.0 bits of final state, and the files grow by 13.1\% with our 16\x2, 13.6\% with our 64\x2 and 3.6\% with our 512\x4 (in the dense mode, whose lanes of 512 values are already 4 times as many, the same count would cost about 24\%). With as many lanes as ours instead, the longest chain would grow 34-fold, beyond that of a single warp (\cref{S:tab:lanes}), and the 16\x2's entropy decoding by about 44\,ms at 0.65\,$\mu$s per value.

\begin{table*}[t]
\centering
\footnotesize
\setlength{\tabcolsep}{4.5pt}
\input{supp/tab_lanes}
\caption{\textbf{Lane length}, on Kodak with one RTX 3090: lanes of about $T$ values give $G$ warps of 32 lanes and, on average, the given number of values per lane; bits per subpixel, and milliseconds per photo of the decoder's GPU part (dec.) and, in a second decode that waits after each kernel, of entropy decoding alone (ent.), each the minimum of 7 rounds after a warm-up. Every decode is exact. Our streams use $T=2{,}048$ for the 16\x2 and 64\x2 and $T=8{,}192$ for the 512\x4 (main paper, Tab.~3).}
\label{S:tab:lanes}
\end{table*}

\subsection{Decoding order}
\label{S:sec:order}
\begin{figure*}[t]
\centering
\includegraphics[width=\linewidth]{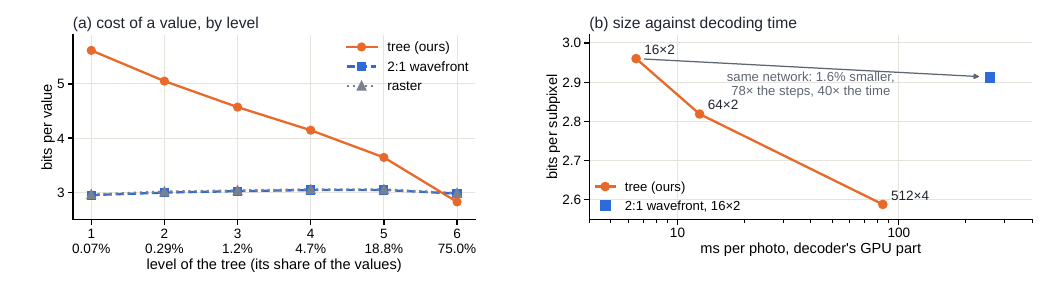}
\vspace{-6mm}
\caption{\textbf{What the order costs and gives.} (a) Bits per value at each level of the tree, on Kodak, for our 16\x2 and for 16\x2 networks trained the same way for a 2:1 wavefront and for raster order (generic networks; each value counted at the level its pixel has in the tree). (b) Size against the GPU part of the decode on one RTX 3090, with the same encoder, lanes and kernels: our 16\x2, 64\x2 and 512\x4, and the wavefront's 16\x2.}
\label{S:fig:order}
\end{figure*}

To separate what the order gives from what our networks, lanes and kernels give, we train our 16\x2 again for two sequential orders, with the recipe of the main paper's 16\x2 (60,000 steps at a peak learning rate of $2\cdot10^{-3}$, then 120,000 at $10^{-3}$, on the same crops) on the RTX 3090. In a \emph{2:1 wavefront}, step $t$ of a channel decodes every pixel with $x+2y=t$, and each value can see every pixel with a smaller $x+2y$: those to its left, those of the row above up to the one above and to its right, and more of the rows further up. No order in which every value sees its left, upper and upper-right neighbours has fewer steps. A landscape Kodak photo takes 5,367 of them, a portrait one 6,135. In \emph{raster} order, every value is a step. Both orders keep the root grid and the order of the channels, and give each value the inputs it has in the tree, read from its 48 nearest neighbours decoded before it, with the inputs that name the level and the step held constant. \cref{S:tab:order} compares them with our 16\x2 under one protocol: the ideal code length of each generic network, and after 300 steps of tuning its output layer to each photo; and, for the wavefront, real files from the paper's encoder (top four colour transforms, 8 orientations, per-photo tuning, lanes of 2,048 values), decoded by the paper's GPU decoder, whose kernels run a step of the wavefront as they run a step of the tree.
The sequential orders make files smaller, but by little: by 2.2--2.5\% with the generic networks, 0.7--1.7\% after tuning, and 1.6\% in the wavefront's real files, from 3.8\% smaller to 1.1\% larger per photo. A 64\x2 trained for the wavefront gains more: 3.0\% with the generic networks and 2.4\% after tuning, against our 64\x2 (3.3 and 2.8\% after the first training cycle of both). The gap does not keep widening with capacity: a 128\x2 trained for the wavefront with the recipe of our 128\x2 gains 2.7\% with the generic networks and 1.1\% after tuning, and a 512\x4, trained in both orders with one matched short recipe (30,000 steps from scratch on the RTX 3090, on the grid of DIV2K crops alone), 1.7\% with the generic networks and nothing after tuning: tuned, the tree's 512\x4 is 0.5\% smaller (2.713 against 2.726 \bpsp; \cref{S:tab:order}): tuning to each photo saves 4.9\% in the tree and 2.7\% in the wavefront. A second training of the tree's first cycle, with the recipe of the other orders, gives code lengths within 0.05\% of our 16\x2's first cycle (3.087 and 2.955 against 3.086 and 2.954 \bpsp, generic and tuned), so the differences between the orders are not those between training runs. They pay about 3.0 bits for a value at every level, while the tree pays 5.6 at its first level and 3.6 at its fifth, but 2.82 at its last, which holds three quarters of the values (\cref{S:fig:order}a).

The wavefront's steps hold about 210 values each on average (220 in a landscape photo), against up to 98,304 in the tree. A Kodak photo takes 5,591 of them on average, 78 times as many as the tree (7 of the 24 streams are in a portrait orientation, which the encoder may choose for any photo), and their number grows with the image's sides: about 24,000 for a 4K frame, where the tree still takes 72. With our lanes and kernels, which we built for a few large steps, the wavefront decodes 40 times slower: 258 instead of 6.5\,ms (\cref{S:fig:order}b), 46\,$\mu$s per step; replaying the whole decode as one CUDA graph, which removes the launches from the host, still takes 235 against 5.2\,ms. A decoder built for many small steps would pay far less per step. On the RTX 3090, a step without work costs 1.2\,$\mu$s as a kernel in a CUDA graph and 0.76\,$\mu$s as a grid-wide barrier in one persistent kernel, over 5,367 steps (0.09\,$\mu$s as a barrier inside one block, on one multiprocessor). Synchronising the wavefront's steps would thus take 4.1--6.2\,ms for a landscape Kodak photo, as long as our 16\x2's whole decode, and 19--28\,ms for a 4K frame, which our 16\x2 decodes in 33\,ms with our kernels; each step would also wait for the prediction and the rANS decoding of its values. Decoding the three channels as a pipeline, which both orders allow, would shorten the wavefront to about 1,800 steps and the tree to about 26. We have not built such a decoder, so the 40\x\ is that of our kernels, not a bound: for small networks, what the order gives is mainly a number of steps that does not grow with the image. With many photos in flight, the wavefront's small steps could also fill the GPU at full load; what the tree keeps is a single photo's time and the few photos needed to fill the GPU. Encoding takes about as long with either order (5.4 and 5.7\,s per photo in this run, which did not cool the GPU as \cref{S:tab:encode} does), and the per-photo update takes the same share of the file.

\begin{figure*}[t]
\centering
\includegraphics[width=\linewidth]{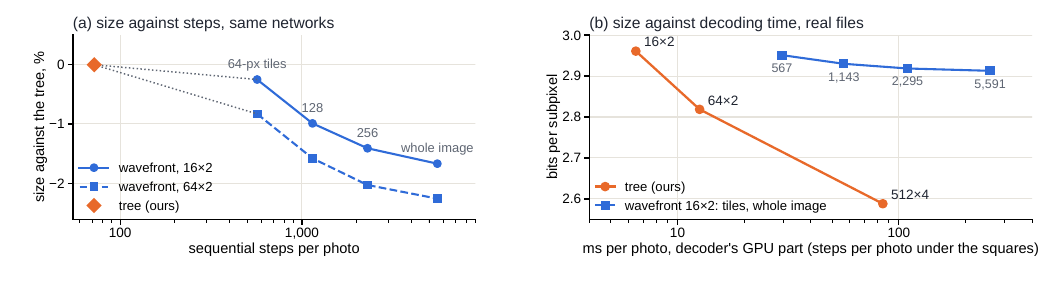}
\vspace{-6mm}
\caption{\textbf{Tiles bound the steps, not the time}, Kodak, one RTX 3090. (a) Ideal code length after tuning against the steps of a photo (main paper, Fig.~5), relative to the tree with the same network: the wavefront networks of \cref{S:tab:order} over the whole image and inside tiles of 64, 128 and 256 pixels (one session). (b) Real files against the GPU part of the decode, with the paper's encoder, lanes and kernels: our 16\x2, 64\x2 and 512\x4 on the tree, and the wavefront 16\x2 inside tiles and over the whole image (the steps of a photo under its squares); every decode is exact.}
\label{S:fig:tiles}
\end{figure*}

\paragraph{Tiles.}
Tiles bound the wavefront's steps too. Inside square tiles of $B$ pixels, all decoded at once, a value sees only the earlier pixels of its tile, the root grid and the earlier colour channels, and a photo takes $3(3B-3)$ steps whatever its size: 567, 1,143 and 2,295 for tiles of 64, 128 and 256 pixels. We run the wavefront networks above in this way, without retraining: a neighbour in another tile is replaced as one outside the image is. The last two blocks of \cref{S:tab:order} give the ideal code lengths, all from one session (the tree's and the whole-image wavefront's differ from the blocks above by at most 0.2\%), and, for the 16\x2, real files and decode times with the paper's encoder, lanes and kernels, every decode exact. Tiles cost the wavefront most of its gain: in tiles of 64 pixels its tuned code length is 1.4\% (16\x2) and 1.5\% (64\x2) above its whole-image value, and only 0.3 and 0.8\% below the tree's; tiles of 256 pixels, with 32 times the tree's steps, leave it 1.4 and 2.0\% below (\cref{S:fig:tiles}a). Real files of the wavefront 16\x2 in 64-pixel tiles are 0.3\% smaller than our 16\x2's but decode 4.6\x\ slower, in 29.6 instead of 6.5\,ms (\cref{S:fig:tiles}b). At every timed wavefront 16\x2 point, tiled or not, our 64\x2 on the tree is both smaller and faster. Networks trained with tile borders might recover part of the loss, but not the context that the borders remove.

\begin{table*}[t]
\centering
\footnotesize
\setlength{\tabcolsep}{5pt}
\input{supp/tab_order}
\caption{\textbf{Decoding order}, our networks on Kodak, one RTX 3090: steps of a photo; ideal code length (the networks' cross-entropy, times 1.005, plus the root and the header; our real untuned files are 0.4--0.7\% above the cross-entropy on Kodak, root, header and the lanes' final states included) of the generic network and after tuning its output layer to each photo (the same for every order); size of the paper's real files; the decoder's GPU part, as the codec launches it and as one CUDA graph (minimum of 7 rounds after a warm-up). Percentages: against the tree with the same network (in the last two blocks, from the same session; the wavefront networks inside tiles without retraining). Every decode is exact.}
\label{S:tab:order}
\end{table*}

\subsection{Root spacing}
\label{S:sec:root}
The spacing $S$ of the root sets the number of levels, and so of steps: a root every 32, 64 or 128 pixels gives 5, 6 or 7 levels and 60, 72 or 84 steps. We train the first cycle of our 16\x2's recipe (60,000 steps) with a root every 32 and every 128 pixels and compare both with our 16\x2's first cycle (\cref{S:tab:root}). The size barely depends on the spacing. With $S=32$ the root, stored raw, costs 0.006 \bpsp more than with $S=64$, which is the whole of the generic network's difference, and 0.5\% after tuning; with $S=128$ the coarse levels see farther-away neighbours, but they hold too few values to matter, and the code length stays within 0.2\% of $S=64$ for 12 more steps.

\begin{table}[t]
\centering
\footnotesize
\setlength{\tabcolsep}{4pt}
\input{supp/tab_root}
\caption{\textbf{Root spacing}, the first training cycle of our 16\x2 on Kodak: levels, steps, the share of the pixels in the root, and the ideal code length (as in \cref{S:tab:order}) of the generic network and after 300 steps of tuning its output layer. $S=64$: our 16\x2's first cycle; $S=32$ and 128 trained the same way on the RTX 3090.}
\label{S:tab:root}
\end{table}

\subsection{Network size}
\label{S:sec:width}
Width stops paying at the top of the family: our 768\x4 makes files as large as the 512\x4 on Kodak (2.589 \bpsp each) and only 0.1--0.2\% smaller on the large photos (main paper, Tabs.~3 and~5), for 1.4--1.6\x\ its decode time. Both read the same inputs and are trained on the same crops, so at this size the data and the inputs, rather than the width, may bound the size of the files (the 768\x4 is the 512\x4 widened and trained for one more cycle, so the two are not fully separated); the 512\x4 is the better trade-off, and we report the 768\x4 to show where the family saturates with our training.

\subsection{Where the bits go}
\label{S:sec:bits}
\begin{figure*}[t]
\centering
\includegraphics[width=\linewidth]{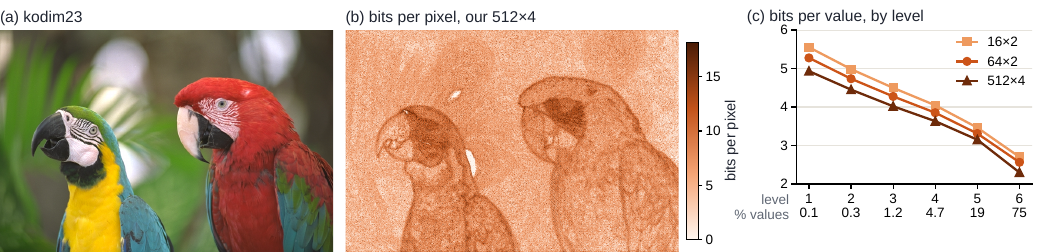}
\vspace{-6mm}
\caption{\textbf{Where the bits go.} (a) A Kodak photo; (b) the bits our 512\x4 spends on each of its pixels (three subpixels; 6.67 bits per pixel, or 2.22 per subpixel, on average, for the coded values alone, without the side information, lane states and header of the file); (c) the bits per value at each of the six levels, over the 24 Kodak photos, with the share of the values that each level holds.}
\label{S:fig:bitsmap}
\end{figure*}

\cref{S:fig:bitsmap} shows the bits our 512\x4 spends on each pixel of a Kodak photo and the cost of a value at each level of the tree. Bits follow image activity: ranked by the Sobel gradient of their brightness (BT.601 luma, averaged over $5\times5$ pixels), the smoothest quarter of the pixels costs 4.8 bits per pixel and the busiest twentieth 10.9. A value of the first level, whose nearest decoded neighbours are up to 45 pixels away, costs about twice as much as one of the last, but the first three levels hold only 1.5\% of the values (main paper, Sec.~3.5).

\section{Encoding and per-photo tuning}
\label{S:sec:encoding}

\subsection{Encoder choices}
\label{S:sec:encchoices}
\cref{S:tab:ablation} adds the encoder's choices of the main paper (Sec.~3.3) one at a time, without changing the networks: the colour transform its fast estimate ranks first, then the best of the top four, the best of 8 orientations, and finally per-photo tuning, whose update the file includes. Tuning gives most of the gain, and more so for larger networks: 3.7\% for the 16\x2, 4.8\% for the 64\x2 and 6.7\% for the 512\x4, against 0.3--0.7\% for the colour transforms and orientations together. The tuned runs keep the transform and the orientation of the untuned ones, so their gain is that of tuning alone. Large photos gain less: on the 11 CLIC.p photos of the main paper's Tab.~3, tuning takes our 512\x4 from 2.263 to 2.238 \bpsp on the RTX 3090 ($-1.1$\%; 2.239 in that table, encoded on the RTX 4090), with an update of 0.14\% of the file.

\begin{table}[t]
\centering
\footnotesize
\setlength{\tabcolsep}{2.4pt}
\input{supp/tab_ablation}
\caption{\textbf{The encoder's choices}, added one at a time, on Kodak: bits per subpixel and the change from the row
above. Lanes as in \cref{S:tab:lanes}.}
\label{S:tab:ablation}
\end{table}

\subsection{Without tuning}
\label{S:sec:untuned}
Where encoding must be faster, the networks can be used without per-photo tuning, like the codecs that use one model for all images. \cref{S:tab:untuned} gives every network's files without tuning, the encoder's other choices kept, against the tuned files of the main paper's Tab.~3. On Kodak, tuning saves 3.7\% for the 16\x2 and more for larger networks, 6.5--6.7\% for the 512\x4 and 768\x4; on the large photos it saves less, and less for larger networks: from 1.8\% (16\x2) to 1.0\% (768\x4) on CLIC.p and from 1.6\% to 1.2\% on DIV2K. Untuned, the 768\x4 is 0.2\% smaller than the 512\x4 on all three sets, a gap that tuning closes on Kodak. Untuned files are decoded by the same kernels with the same work, without merging an update. Their encoder still searches the colour transforms and orientations, coding each photo 11 times: 1.8--2.4\,s per Kodak photo from the 16\x2 to the 512\x4 (the first 8 photos, each after the GPU has cooled to 60\,\textdegree C), and, without the cooling, 22\,s per CLIC.p photo for the 512\x4; with the first-ranked transform and orientation only, 1.1--1.4\,s per Kodak photo, for files 0.3--0.7\% larger (\cref{S:tab:ablation}).

\begin{table*}[t]
\centering
\footnotesize
\setlength{\tabcolsep}{5pt}
\input{supp/tab_untuned}
\caption{\textbf{Every network without per-photo tuning}, bits per subpixel, against the tuned files of the main paper's Tab.~3 (top four colour transforms, 8 orientations and lanes as there; gain: what tuning saves, update included). Untuned files encoded on the RTX 3090 (the main paper's Tab.~4 gives 2.7755 for the 512\x4, encoded on the RTX 4090; \cref{S:sec:exact}). Every stream is decoded and checked.}
\label{S:tab:untuned}
\end{table*}

\subsection{Encoding time}
\label{S:sec:enctime}
\cref{S:tab:encode} gives the size and the encoding time with 200 and with 25 tuning iterations. The time covers the encoder's choices, the tuning and the probabilities of every value; we time 8 photos, each after the GPU has cooled to 50\,\textdegree C, so that the temperature limit of our machine never paused a timed encode. With 25 iterations instead of 200, files grow by 1.1\% (16\x2), 1.7\% (64\x2) and 3.4\% (512\x4), and encoding is 1.7, 1.9 and 3.4\x\ faster: 2.6--4.4\,s instead of 4.5--14.9\,s per photo. Without the cooling, the same encodes took 4.5, 4.9 and 22.2\,s per photo, the 512\x4 heating the GPU until our machine's temperature guard paused it. On large photos, encoding takes several times longer: for the CLIC.p photos of the main paper, our encodes on the RTX 4090 took 32--51\,s per photo on average, depending on the network. \cref{S:tab:encbreak} shows where this time goes on the RTX 3090: tuning takes only 2.5--2.7\,s for the 16\x2 and 64\x2 and 12\,s for the 512\x4 of their 29--46\,s. Most of the rest went to three host-side parts of that encoder: the fast estimate that ranks the 42 colour transforms (11\,s), the index tables of the photo's two orientations' sizes (8--9\,s) and the choice of the update's quantisation (4--6\,s).

\subsection{A faster encoder}
\label{S:sec:fastenc}
Our faster encoder moves these parts to the GPU (\cref{S:tab:encfast}). A first version changes no byte: it ranks the colour transforms from residual histograms computed in one kernel (the same four best transforms for every photo), builds the index tables in one kernel (\cref{S:tab:cold}) and copies all frequencies to the host at once. It writes the files of the main paper's encoder byte for byte for every photo we tried: 24 Kodak photos with the 16\x2 and 64\x2 and 8 with the 512\x4, untuned and tuned, and 4 large photos with the 16\x2, 64\x2 and 512\x4, tuned; in 0.04--0.06\,s per Kodak photo untuned with the small networks (first-ranked colour transform, identity orientation) instead of 1.3--1.4\,s, and tuned in 10.5--12.2\,s per large photo instead of 29--33\,s. The 512\x4's long tuned encodes heat the RTX 3090 until our machine's temperature guard pauses them, unless they start cold (as in \cref{S:tab:encode}, which drops paused timings), so we time the 512\x4 on the A100, and on the RTX 3090 only the pool encoder below, cooled to 45\,\textdegree C before each encode, with none of its timed encodes paused. A second version also scores every candidate (colour transform, orientation and quantisation step of the update) by its exact code length under the decoder's integer distributions, instead of the training-time estimate, and runs the 200 tuning steps as one captured CUDA graph, with the same loss, optimiser and learning-rate schedule. With the 16\x2 and 64\x2 its files are within 0.05\% of the main paper's, in 2.3--2.9\,s per Kodak photo and 3.1--4.2\,s per large photo (on an A100: 2.1\,s per Kodak photo with the 16\x2, 2.5--3.3\,s per large photo). On the A100, the 512\x4 takes 8.9\,s per Kodak photo and 13--14\,s per large photo (40--43\,s with the main paper's encoder), 87\% and 59--61\% of it in tuning. Tuned on a pool of 262,144 values whose features are computed once, in batches of 16,384 values instead of 65,536, it takes 3.5\,s per Kodak photo and 7.2--7.8\,s per large photo, with files 0.15\% and 0.07--0.09\% larger, and the 768\x4 4.6\,s per Kodak photo (0.11\% larger); on the RTX 3090, the 512\x4 takes 5.0 and 12.9--14.4\,s, with files 0.14\% and 0.05\% larger, the 16\x2 0.85 and 2.0\,s and the 64\x2 1.1 and 2.6\,s, with files 0.01--0.16\% larger. On large photos, 57--58\% of its time on the A100 goes to the exact code lengths of its candidates, each a full pass of the 512\x4 over the photo, and 31--32\% to tuning (67\% on Kodak). For comparison, on the A100, LC-FDNet takes 1.1\,s to encode a Kodak photo and 8.0\,s for a DIV2K photo, and ArIB-BPS 3.8 and 28.9\,s.

\begin{table*}[t]
\centering
\footnotesize
\setlength{\tabcolsep}{4pt}
\input{supp/tab_encfast}
\caption{\textbf{A faster encoder}, seconds per photo: Kodak (24 photos), each after a warm-up of its size, and 2K photos (RTX 3090: two CLIC.p and two DIV2K photos of the main paper's Tab.~3; A100: DIV2K 0801 and 0805), each its size's first encode, tables included; every timed encode starts after the GPU has cooled to 55\,\textdegree C. Untuned: first-ranked colour transform, identity orientation; tuned: as in the main paper (top four transforms, 8 orientations, 200 tuning steps). Paper: the main paper's encoder; same: its host steps on the GPU, the same bytes; fast: exact code lengths for every choice and the tuning as one CUDA graph; pool: tuning on 262,144 precomputed values in batches of 16,384. In parentheses: the change in file size against the paper's encoder on the same photos and GPU (A100, Kodak: its streams of \cref{S:tab:a100}). RTX 3090, 512\x4: only the pool encoder, cooled to 45\,\textdegree C before each encode, none of whose timed encodes our machine's temperature guard paused; it paused the others' long encodes. Every stream is decoded and checked.}
\label{S:tab:encfast}
\end{table*}

\begin{table}[t]
\centering
\footnotesize
\setlength{\tabcolsep}{1.9pt}
\input{supp/tab_encbreak}
\caption{\textbf{Where the encoder's time goes} on large photos, seconds per photo on one RTX 3090, for two of the CLIC.p and DIV2K photos of the main paper's Tab.~3, each encoded after the GPU has cooled to 50\,\textdegree C: ranking the colour transforms, building the index tables, coding with the top four transforms and in the other orientations, tuning (200 iterations), choosing the update's quantisation, and computing and packing the probabilities.}
\label{S:tab:encbreak}
\end{table}

\subsection{Small images}
\label{S:sec:small}
The per-photo update costs side information, a larger share of a small file. On centre crops of the Kodak photos (\cref{S:tab:thumbs}), tuning still makes files smaller: by 0.9--1.6\% for crops of $64\times64$ pixels, 2.3--2.8\% for $128\times128$ and 3.4--4.0\% for $256\times256$, against 3.7--6.7\% for the whole photos. The encoder keeps a colour channel's update only where it shortens the file, so on the smallest crops it drops it for 11--44\% of the channels; the updates it keeps take 1.7--2.1\% of those files.

One image at a time, a small image still takes the tree's 72 steps (\cref{S:tab:thumbtime}). Our 16\x2 decodes a $64\times64$ crop in 3.2\,ms and a $256\times256$ one in 5.5\,ms on the RTX 3090, where PNG, WebP and JPEG XL take 0.06--0.5\,ms and 0.8--10.7\,ms on one thread of its CPU. From $256\times256$ pixels on, our 64\x2 matches the size of JPEG XL at effort 9 in less time (8.1 against 10.7\,ms); below that, the CPU codecs are faster. Untuned, our 64\x2 and 512\x4 are smaller than every CPU codec at every crop size, by 0.1--3.0\% and 6.0--8.5\% against JPEG XL e9, and our 16\x2 is 0.8--3.9\% larger than it. The first decode of a size that the process has not seen, which builds the size's index tables, takes 14--118\,ms for these crops.

\begin{table}[t]
\centering
\footnotesize
\setlength{\tabcolsep}{3.2pt}
\input{supp/tab_thumbtime}
\caption{\textbf{Small images, one at a time}: centre crops of the 24 Kodak photos; bits per subpixel and decode time per crop, our networks untuned on the RTX 3090 (end to end, minimum of 7 rounds after a warm-up) and the CPU codecs with one thread of its i9-14900K. Every decode is exact.}
\label{S:tab:thumbtime}
\end{table}

\begin{table}[t]
\centering
\footnotesize
\setlength{\tabcolsep}{3.4pt}
\input{supp/tab_thumbs}
\caption{\textbf{Per-photo tuning on small images}: centre crops of the 24 Kodak photos, bits per subpixel untuned and tuned as in the main paper; kept: share of the colour channels whose update the encoder keeps; update: its share of the tuned file. Every stream is decoded and checked.}
\label{S:tab:thumbs}
\end{table}

\begin{table}[t]
\centering
\footnotesize
\setlength{\tabcolsep}{3pt}
\input{supp/tab_encode}
\caption{\textbf{Encoding with fewer tuning iterations}, on Kodak with one RTX 3090: bits per subpixel over the 24 photos, and seconds per photo over the first 8, each timed after the GPU has cooled to 50\,\textdegree C.}
\label{S:tab:encode}
\end{table}

\section{Exact decoding on any GPU}
\label{S:sec:exact}

\subsection{Integer probabilities}
\label{S:sec:intprob}
The encoder and the decoder turn a network's mixture into the integer cumulative distribution $F$ of the main paper (Sec.~3.4) by the same steps. The mixture weights, from a softmax, are cut to integers, $w_k=\lfloor 4096\,\pi_k\rfloor$, and the largest component takes the remainder, so that they sum to 4,096. Each logistic has its mean $\mu_k$ and a log-scale clamped to $[-6,6]$; its argument $2^{16}(v-\tfrac12-\mu_k)/s_k$ is one fused multiply--add in fp32, $v\,a_k+c_k$, clamped to $[-2^{20},2^{20}-1]$. A table of 8,193 entries, $\mathrm{round}(2^{16}\sigma((i-4096)/256))$, whose ends are exactly 0 and $2^{16}$, gives the sigmoid, interpolated linearly in the low 8 bits of the argument with integers. The weighted sum $C(v)=\sum_k w_k\sigma_k$ is at most $2^{28}$, and $F(v)$ is 0 for $v\le v_0$, $2^{16}$ above the range, and $(v-v_0)+\lfloor C(v)(2^{16}-n)/2^{28}\rfloor$ in between, from the high half of a 64-bit product. Every value of the range thus keeps a frequency of at least 1 in $2^{16}$.

\subsection{Portability}
\label{S:sec:portability}
\cref{S:tab:xgpu} decodes on the RTX 3090, end to end as in the main paper, the streams that we encoded on the RTX 4090 (CUDA 12.8, PyTorch 2.8) for the main paper. The fast and dense modes and the 16\x2 to 64\x2 networks decode them bit for bit; the 128\x2, 512\x4 and 768\x4, which run their layers as matrix multiplications in cuBLAS, do not: cuBLAS guarantees identical results only on GPUs of the same architecture and number of multiprocessors, with the same library version and while a single stream is active~\cite{cublas}, because it picks its algorithm for a matrix product, and with it the order of the sums, such as how the inner dimension is split between thread blocks, by the GPU. A second pair agrees as well: an A100 (CUDA 12.6, PyTorch 2.7.1), another chip than both (compute capability 8.0 against 8.6 and 8.9), writes the same untuned streams as the RTX 3090 (identity orientation, first-ranked colour transform), byte for byte, for every Kodak photo and each of the 16\x2, 32\x2 and 64\x2, from the same models and code: the two GPUs compute every value's probability alike, so each decodes the other's streams. For the 128\x2, 512\x4 and 768\x4, none of the 24 streams agree. These are two pairs of GPUs, each with the GPU and the software changed together, and our kernels for the small networks use the GPU's approximate exponential, which other GPU generations need not compute alike; our 8-bit CPU networks use integer layers, a polynomial activation and an exponential of their own instead, which any processor that follows IEEE 754 computes alike. The fast and dense modes need no such care: their decoders compute with integers only, and our CPU and GPU decoders of the fast mode read the same files.

\subsection{Portable kernels}
\label{S:sec:portable}
A variant of our kernels removes both causes (\cref{S:tab:port}). It replaces the approximate exponential, in the SiLU and in the mixture's weights and scales, by one of its own (Cody--Waite reduction and a degree-6 polynomial in fused multiply--adds, within about one ulp) and the one logarithm, of an integer from 1 to 511, by a table; and it runs the layers of the 128\x2 to 768\x4 in fp32, in a kernel of its own in which every output is the bias plus the products summed in increasing order of the inner index, one fused multiply--add each, whatever the tiling. Every result is then defined by correctly rounded IEEE-754 operations in a fixed order. The encoder and the decoder both use these kernels, so the streams change, by at most 0.005\% in size. The RTX 3090 and the A100 then write the same streams, byte for byte, for every Kodak photo and each of the six networks, the 128\x2, 512\x4 and 768\x4 included, whose standard streams differ on every photo; each GPU decodes them exactly. On the A100, the 16\x2 to 64\x2 decode 3--5\% slower, the 128\x2 6\% faster, and the 512\x4 and 768\x4, whose layers no longer run on the tensor cores, 3.0 and 4.4\x\ slower; on the RTX 3090, whose tensor cores are slower relative to its fp32 units, the 512\x4 decodes 1.7\x\ slower. Our layer kernel reaches 17.7--18.2 of the A100's 19.5 fp32 TFLOPS, so the loss is that of fp32 arithmetic against bf16 tensor cores: a layer of 512 neurons over 262,144 values takes 7.6\,ms against 1.6\,ms in cuBLAS (RTX 3090: 5.9 against 3.4\,ms, at 23.3 of its 35.6 fp32 TFLOPS).

\subsection{Portable at tensor-core speed}
\label{S:sec:int8}
Sums of integers are exact in any order, so a product of 8-bit integer matrices gives the same bits on every GPU, whatever kernel computes it. We run the hidden layers of the 512\x4 and 768\x4 so, on the integer tensor cores (\cref{S:tab:port8}), without retraining the networks: each row of activations entering a hidden layer is quantised to 8 bits with its own scale, its largest magnitude over 127, and each output channel of the weights likewise, once, on the host; the tensor cores sum the products in 32-bit integers, and one kernel per row converts the sums to fp32, scales them back, adds the bias and applies the portable SiLU, in correctly rounded IEEE-754 operations, then quantises the result for the next layer. A variant quantises each row over its range, with a zero point, since SiLU's outputs reach only $-0.28$ below zero; its sums gain an exact integer correction. The first layer, whose inputs mix scales, and the head stay in the portable fp32 kernel. The RTX 3090 and the A100 write the same integer streams, byte for byte, for every Kodak photo, and each decodes them exactly. Quantising networks trained in floating point makes the files 0.11--0.16\% larger with a zero point, 0.20--0.25\% without; networks trained with 8-bit layers, as our CPU networks are, could lose less. Tuned to each photo, as are the streams of the main paper, the files would be 0.43 and 0.29\% larger with a zero point if the update were tuned in floating point and quantised afterwards. Our encoder therefore tunes it through the integer layers: its forward pass computes them exactly, from rows quantised as the decoder quantises them, and passes the gradients straight through the rounding. The tuned files are then 0.25 and 0.21\% larger on Kodak (\cref{S:tab:port8}), 0.09 and 0.08\% on CLIC.p/4 and 0.12 and 0.11\% on DIV2K/4. On the RTX 3090, the integer kernels decode as fast as the standard ones: the 512\x4 in 85.3 against 84.6\,ms and the 768\x4 in 125 against 130\,ms, against 143 and 267\,ms with the fp32 portable kernels. Tuned, on the RTX 3090, they decode a Kodak photo in 91.7 against 89.4\,ms and 131.8 against 134.8\,ms, the CLIC.p/4 photos in 429 against 442\,ms and 622 against 721\,ms, and the DIV2K/4 photos in 504 against 513\,ms and 729 against 839\,ms, every decode exact. An RTX 4090 decodes the 48 tuned integer streams of Kodak that the RTX 3090 wrote, all exactly, in 64.7 and 93.3\,ms, against 53.6 and 76.9\,ms with its standard kernels on streams it wrote itself (2.5884 and 2.5897 \bpsp, so the integer files are 0.25 and 0.21\% larger there too): 1.2\x\ the time. On the A100, in one session, the 512\x4 decodes in 89\,ms against 81 and 168\,ms, and the 768\x4 in 127 against 86 and 312\,ms; there the host limited the standard decodes: sampled every 100\,ms, the GPU was busy 57 and 73\% of their time, against 98\% of the fp32 portable decodes' and 75--76\% of the integer ones' (in the session of \cref{S:tab:port}, the standard decodes took 55 and 71\,ms).

\begin{table}[t]
\centering
\footnotesize
\setlength{\tabcolsep}{2.8pt}
\input{supp/tab_port}
\caption{\textbf{Portable kernels} on Kodak: untuned streams (first-ranked colour transform, identity orientation), the size of the portable streams against the standard ones, the decode time per photo on the A100 with the standard and the portable kernels (end to end, best of 3 warm rounds, mean over the 24 photos) and their ratio, and how many of the 24 streams the A100 and the RTX 3090 write byte for byte alike, with the standard and with the portable kernels. Every decode is exact. A first, simpler layer kernel, with the same results, made the 512\x4 and 768\x4 3.8 and 5.6\x\ slower.}
\label{S:tab:port}
\end{table}

\begin{table}[t]
\centering
\footnotesize
\setlength{\tabcolsep}{2.4pt}
\input{supp/tab_port8}
\caption{\textbf{Portable at tensor-core speed}: the 512\x4 and 768\x4 on Kodak with the standard kernels, the portable fp32 kernels of \cref{S:tab:port}, and portable kernels whose hidden layers run in 8-bit integers (symmetric or asymmetric rows): size against the standard streams, decode time per photo (end to end, best of 3 warm rounds, mean over the 24 photos; one session per GPU; tuned rows: minimum of 7 rounds after a warm-up, the two kernels in two sessions) and its ratio to the standard kernels', and how many of the 24 streams the two GPUs write byte for byte alike. Untuned streams but for the tuned rows, lanes of 8,192. Every decode is exact. RTX 3090: the 768\x4's runs each started with the GPU cooled to 45\,$^\circ$C.}
\label{S:tab:port8}
\end{table}

\begin{table}[t]
\centering
\footnotesize
\input{supp/tab_xgpu}
\caption{\textbf{Streams of the RTX 4090, decoded on an RTX 3090}, Kodak: exact or not, and the end-to-end decode time per photo on the RTX 3090 (minimum of 2 rounds), whose host CPU is faster than that of our server. A separate session from \cref{S:tab:lanes}: times differ between sessions by up to about 20\%, so these end-to-end times can fall below the GPU parts timed there.}
\label{S:tab:xgpu}
\end{table}

\section{Measurement notes}
\label{S:sec:measure}

\subsection{Timing protocol}
\label{S:sec:protocol}
\cref{S:tab:protocol} lists, for each group of results, the machine, the software, what the clock covers and which statistic we report. Every timed decode is checked against the original, bit for bit. The statistics differ between groups: in the main paper, our times are minima of 7 rounds, those of the learned codecs we ran on the A100 minima of 3 (ArIB-BPS: one round, timed by its own evaluation), and DLPR's the mean of single decodes without the first image; the runs of this supplement use the rounds stated with each table (2 to 7). More rounds can only lower a minimum, and medians, at most 19\% above the minima, turn no comparison against \fmt (\cref{S:sec:everycodec}).

\begin{table*}[t]
\centering
\footnotesize
\setlength{\tabcolsep}{3pt}
\begin{tabular}{@{}>{\raggedright\arraybackslash}p{3.1cm}>{\raggedright\arraybackslash}p{3.2cm}>{\raggedright\arraybackslash}p{2.5cm}>{\raggedright\arraybackslash}p{4.3cm}>{\raggedright\arraybackslash}p{3.3cm}@{}}
\toprule
Results & Machine & Software & Clock & Statistic \\
\midrule
Main paper: Fig.~4 and Tab.~3, one image at a time & RTX 4090; host AMD EPYC 7702 & CUDA 12.8, PyTorch 2.8 &
from the compressed bytes in host memory to the RGB image in GPU memory: parsing, merge of the update and inverse
colour transform inside; index tables of the image size built once, outside & minimum of 7 rounds after a warm-up,
mean over the images \\
CPU codecs (main paper and \cref{S:sec:everycodec}) & Core i9-14900K, 32 threads, WSL2 & imagecodecs 2026.3.6 (libjxl
0.11.2, libwebp 1.6.0, \ldots) & from the compressed bytes to the image, both in host memory & as above, at each
codec's best number of threads \\
Throughput (main paper, Tab.~6; \cref{S:sec:throughput}) & as above & as above & many images at once, batches on
several host threads & best number of threads and batch size; megapixels per second \\
Published times (main paper, Tab.~4 and Fig.~4) & RTX 4090D (HPAC's measurements); \mbox{CALLIC}: not stated & --- &
as stated by the authors & as published \\
A100 runs (main paper, Tab.~4$^\ddagger$; \cref{S:tab:a100,S:tab:port}) & A100-SXM4-40GB in a shared server & CUDA 12.6,
PyTorch 2.7.1; rivals in their own environments & ours as in the first row; rivals from a bitstream in memory or on
a RAM disk to the decoded image & ours: minimum of 7 rounds (\cref{S:tab:port,S:tab:port8}: best of 3); rivals: minimum of 3 after a checked decode
(ArIB-BPS: one round, CUDA events) \\
RTX 3090 runs (\cref{S:sec:design,S:sec:encoding,S:sec:exact,S:sec:memsize}) & RTX 3090; host Core
i9-14900K, WSL2 & CUDA 13.0, PyTorch 2.11 & as in the first row, but the lane runs time the decoder's GPU part, from
the parsed stream to the image; encoders: from the image in host memory to the stream in host memory & minimum of
the stated rounds; a GPU temperature guard (pause above 73\,$^\circ$C) is
logged, and no timing that it paused is used (except the 33-MP 512\x4 decodes of \cref{S:tab:large}, an upper bound) \\
\bottomrule
\end{tabular}
\vspace{-2mm}
\caption{\textbf{Timing protocol} of every group of results. Times from different sessions on the same GPU differ by up to about 20\%, so we compare only times of the same session, except our portable kernels against FNLIC on the A100 (main paper, Sec.~4.2), whose 3.1\x\ is far beyond that spread.}
\label{S:tab:protocol}
\vspace{-3mm}
\end{table*}

\subsection{DLPR}
\label{S:sec:dlpr}
We ran DLPR with its authors' code and weights, in its original software stack (PyTorch 1.11, CUDA 11.3), with one function, erfc, replaced by $1-\mathrm{erf}$ so that it compiles for our GPU. Its decode took 5.37\,s per Kodak photo; it decodes its residuals with an arithmetic coder on our server's host CPU, one value after another; we report the mean of single decodes, without the first image, and compare with its published time in the main paper.

\subsection{Averages over a set}
\label{S:sec:averages}
The main paper gives all bits over all subpixels of a set, the convention of the published tables: on CLIC.p, their JPEG XL and JPEG 2000 rows match our runs of these codecs under it, and not the mean of the photos' bits per subpixel (\cref{S:tab:means}); on DIV2K the two are within rounding for JPEG XL. That mean, which weighs every photo alike whatever its size, is 2.3--2.6\% higher on CLIC.p and 0.3\% on DIV2K, for every codec alike, so it changes no comparison; on Kodak, whose photos all have the same size, the two coincide.

\begin{table}[t]
\centering
\footnotesize
\setlength{\tabcolsep}{2.2pt}
\input{supp/tab_means}
\caption{\textbf{Two averages of the full validation sets}, bits per subpixel: all bits over all subpixels of the set (main paper, Tab.~5), the mean over photos, and the published values of HPAC's Tab.~1~\cite{hpac}. Our networks tuned to each photo; HTJ2K and JPEG 2000 by nvJPEG2000 (per photo: not cached for DIV2K).}
\label{S:tab:means}
\end{table}

\subsection{Photo by photo}
\label{S:sec:paired}
Kodak has only 24 photos, so we check the main paper's size comparisons photo by photo (\cref{S:tab:paired}): for each pair, the difference over the set, a 95\% bootstrap interval over its photos, and on how many photos ours is smaller. The comparisons the main paper draws hold on most photos, with intervals that exclude zero: our tuned 64\x2 is smaller than JPEG XL e9 on 20 of the 24 Kodak photos and on all 25 of DIV2K/4, and smaller than FNLIC on 20 and 19; our tuned 512\x4 is smaller than FNLIC on 24 and 24. Three are ties within the spread of the photos: our untuned 512\x4 against ArIB-BPS on Kodak ($-$0.2\%, smaller on 15 photos), and both of our large networks against ArIB-BPS on DIV2K/4 ($+$1.0 and $+$1.2\%, smaller on 6 of 25 photos, intervals across zero). Untuned, our 64\x2 is 0.8\% larger than LC-FDNet; tuned, 4.1\% smaller, on every photo. The 768\x4 is no smaller than the 512\x4 on Kodak ($+$0.1\%), and smaller on every DIV2K/4 photo ($-$0.2\%).

\begin{table}[t]
\centering
\footnotesize
\setlength{\tabcolsep}{2.6pt}
\input{supp/tab_paired}
\caption{\textbf{Photo by photo}: the size of \fmt against another codec, all bits over all subpixels of the set (negative: ours is smaller), a 95\% bootstrap interval (10,000 resamples of the set's photos), and on how many photos ours is smaller. Ours tuned to each photo, the streams of our A100 runs; $^\circ$: untuned (best of four colour transforms and eight orientations), against codecs with one model for all images. FNLIC, LC-FDNet and ArIB-BPS as in
\cref{S:tab:a100}; JPEG XL by libjxl 0.11.2.}
\label{S:tab:paired}
\end{table}

\subsection{Published times}
\label{S:sec:published}
The published times of learned codecs in the main paper are those that HPAC's authors measured on an RTX 4090D, with about 11\% fewer multiprocessors than our RTX 4090, for every codec of their Tab.~3~\cite{hpac}, except \mbox{CALLIC}'s, which comes from its own paper (1.7\,s per Kodak photo, its Tab.~2; the GPU is not stated)~\cite{callic}. FNLIC's paper gives 13\,ms of ``decoding inference'' per Kodak photo on a GTX 1080 Ti, and its table times LC-FDNet at 0.61\,s, against 1.71\,s in HPAC's measurement; in our runs, below, FNLIC's whole decoder takes 0.53\,s (0.52\,s without deleting its temporary files, the time we compare) and LC-FDNet 1.0\,s. We keep HPAC's measurements for the codecs we did not run, the only ones that cover them all with one protocol. We found no public code for HPAC or \mbox{CALLIC} with which to time them under our protocol; DLPR, which we could run, took 3.8\x\ its published time, and ArIB-BPS, run on an A100 (below), 0.35\x.

\subsection{FNLIC, LC-FDNet and ArIB-BPS on an A100}
\label{S:sec:a100}
We ran them with their authors' code and models on one A100-SXM4-40GB, beside our decoders (\cref{S:tab:a100}), one photo at a time, from the compressed bytes to the pixels: for FNLIC and LC-FDNet the minimum of 3 rounds after a decode that we check against the photo (7 rounds for ours, as in the main paper). FNLIC fits latents and a small network to each photo, in $2\times130{,}000$ iterations; we took the fitted models its authors released, encoded them on the A100 with their encoder, and timed their decoder in its own environment (PyTorch 2.0.1, CUDA 11.7) from a bitstream file on a RAM disk. It takes 535\,ms per photo, 450\,ms of them in the 12 sub-image channels decoded one after another, each by computing a distribution for every value on the GPU, copying them to the host and decoding the channel there with torchac's arithmetic coder (213\,ms in all), and 17\,ms in the shell commands it spawns to delete temporary files; without autograd it takes the same (536\,ms). Even with a free arithmetic coder and without deleting files, it would take 304\,ms, 24\x\ our tuned 64\x2 there: most of its time goes to the 12 channels decoded one after another, each a round trip between the GPU and the host. The bitstreams its authors released, of the same size (2.881 \bpsp), decode wrongly on the A100, all 24, with 87--99.6\% of the pixels wrong, as its documentation warns may happen on another device. LC-FDNet codes a quarter of the pixels with JPEG XL and the rest in 9 steps, each running its network twice on the GPU and torchac twice on the host. We follow its encoder and decoder scripts with its checkpoints trained on Flickr2K, with three changes that can only make it faster: the quarter goes through libjxl inside the process (lossless, effort 7, as its command-line tools at their default) instead of through PNG files, the streams stay in memory, and the GPU's memory cache is not emptied between steps. It reaches 2.938 \bpsp (2.98 in HPAC's table) in 997\,ms per photo: 465\,ms until the distributions reach the host, 220\,ms in torchac and 30\,ms in JPEG XL. ArIB-BPS codes the image bit plane by bit plane with a U-Net; with its released ImageNet64 models and its own evaluation (one round per photo: encode, decode and comparison, timed by CUDA events) it makes 2.781 \bpsp (2.78 in HPAC's table), decodes every photo exactly in 3.59\,s and encodes it in 3.78\,s. Its network needs sides divisible by 16, so on DIV2K/4 we pad each photo on its bottom and right by edge replication, the padding's bits counted; 10 of the 25 photos do not fit whole into the A100's 40\,GB, so we code them in two halves, each padded likewise. It then makes 2.516 \bpsp, 1.2\% less than our 512\x4 and 1.0\% less than our 768\x4, and decodes every photo exactly, in 25.9\,s on average, 106 and 71\x\ their time. Our decoders are slower on the A100 than on the RTX 4090 with the small networks (16\x2: 7.6 against 4.5\,ms) and about as fast with the large ones (512\x4: 58.7 against 53.5\,ms).

\begin{table}[t]
\centering
\footnotesize
\setlength{\tabcolsep}{5pt}
\input{supp/tab_a100}
\caption{\textbf{On one A100}, Kodak and DIV2K/4 (every fourth DIV2K validation photo): bits per subpixel and decode time per photo, one at a time, from the compressed bytes to the pixels, every decode checked; our streams encoded on the A100 (tuned, lanes as in the main paper's Tab.~3), HTJ2K and JPEG 2000 by nvJPEG2000, FNLIC, LC-FDNet and ArIB-BPS with their authors' code and models (FNLIC's whole decoder, with its 17\,ms of deleting temporary files). Every decode is exact but one: LC-FDNet decodes one pixel of DIV2K 0805 wrongly. ArIB-BPS codes 10 of the DIV2K/4 photos in two halves, which do not fit whole into the A100's 40\,GB.}
\label{S:tab:a100}
\end{table}

\subsection{CPU codecs}
\label{S:sec:cpu}
The CPU codecs run under WSL2 on the i9-14900K, whose 8 performance and 16 efficiency cores give 32 threads. We do not pin threads to cores, but give each codec its best number of threads; pinning a single thread to any core gives the same time, as Windows moves it to a performance core.

\subsection{Two dense decoders}
\label{S:sec:twodense}
The dense format has two GPU decoders, one built for a single photo and one for full load, each 1.2--1.5\x\ faster than the other in its own regime; we time each regime with its faster decoder.

\subsection{Reproducibility}
\label{S:sec:repro}
We keep the logs of every measurement and recompute every reported number from them with one script. One CPU point, JPEG-LS on Kodak, is a re-measurement, since the session of its first measurement kept no output.

\subsection{Newer versions of JPEG XL}
\label{S:sec:jxl12}
Every JPEG XL number in the main paper and above is from libjxl 0.11.2, through imagecodecs 2026.3.6. libjxl 0.12 (July 2026) changed its lossless coding and its faster-decoding levels, so we measured again with libjxl 0.12.0, through imagecodecs 2026.11.11, and with libjxl 0.13.0 through its command-line tools: the same protocol on the same CPU, efforts 1 to 9, faster-decoding levels 0 to 4, every decode exact. The two versions write files of the same size. On Kodak, effort 9 makes files 2.0\% smaller than 0.11.2 (2.828 against 2.885 \bpsp) and decodes them in 12.4 instead of 11.0\,ms, so our tuned 64\x2 is 0.4\% smaller than it and 1.7\x\ faster; with faster decoding (level 1: 2.860 \bpsp, 8.5\,ms) it is larger and still slower than our 64\x2. On CLIC.p/4 and DIV2K/4, effort 9 makes files 1.1 and 1.3\% larger than 0.11.2 (2.485 and 2.821 \bpsp), and on the full sets 0.9 and 1.0\% larger (2.650 and 2.816). On all three sets, no setting of either version is both smaller and faster than any \fmt mode; on Kodak, two settings join the frontier, between our dense mode and our 16\x2: effort 9 at faster-decoding level 2 (3.032 \bpsp, 4.05\,ms) and effort 3 at levels 2 and 3 (3.182 \bpsp, 3.3\,ms), as effort 3 does in the main paper. Through libjxl's own decoder, which keeps its threads from one decode to the next, the faster-decoding levels take up to 1.7\x\ less time on Kodak than through imagecodecs (effort 9, level 2: 2.42\,ms), and the same holds. For decoding images into a training job, the comparison that matters is at full load, where JPEG XL leaves its images in host memory and takes the CPU's cores from the rest of the input pipeline (\cref{S:sec:throughput}). We will report both versions in the main paper's tables in its final version.

\section{Memory, image sizes and host work}
\label{S:sec:memsize}

\subsection{Memory}
\label{S:sec:memory}
Our decoder keeps, per image size, its index tables and the buffers of the size's first decode, and reuses them for later images of that size. In the main paper's runs, its parser also built the tables of our CPU decoder, which the GPU decoder does not read: 0.8\,kB per pixel more, 1.1\,kB in all for our 16\x2 and 64\x2, so that a 33-megapixel image needed 33.6\,GB and spilled from the RTX 3090's 24\,GB into host memory. Our released parser (the light parse) takes from them only the number of root pixels; the streams, the kernels and the decode times do not change (30.6 against 30.7\,ms for a 4K frame with our 16\x2 on an A100). Our 16\x2 and 64\x2 then keep 276 bytes per pixel, 228 of them tables, and our 512\x4, with lanes of 8,192 values and its layers in cuBLAS, 311--403 (\cref{S:tab:large}): a 33-megapixel image needs 8.5\,GB with the small networks and 9.6\,GB with the 512\x4, and our 16\x2 decodes it in 133\,ms on the RTX 3090 instead of 725\,ms with the unused CPU tables. Once these are in place, a decode of a Kodak photo by our 16\x2 needs only 22\,MB more. Every entry of the index tables is a function of the value's index alone, which is how one GPU kernel builds them (\cref{S:sec:tables}), so the decoding kernels could compute them on the fly instead, trading memory for index arithmetic in every step; we have not built that variant. The main paper's encoder, untuned, needs 8.4, 15.2 and 27.0\,GB at these sizes, with every network (GB: $2^{30}$ bytes). Our faster encoder (\cref{S:tab:encfast}) needs less, measured above the network's weights on the RTX 3090, every stream decoded and checked: tuned, 1.1\,GB for a Kodak photo with our 16\x2 and 64\x2 and 3.3\,GB with our 512\x4, 4.2 and 6.4\,GB for a 2K photo (DIV2K 0801) and 12.4 and 13.6\,GB for a 4K frame, 1.6--1.8\,kB per pixel there; tuned on the pool, 2.1, 5.2 and 12.4\,GB with each network; untuned, 875 bytes per pixel (512\x4: 0.6, 2.9 and 6.9\,GB). A 24\,GB GPU thus tunes our networks to frames of up to 4K.

\begin{table}[t]
\centering
\footnotesize
\setlength{\tabcolsep}{3pt}
\input{supp/tab_large}
\caption{\textbf{Very large images} on one RTX 3090 (24\,GB): GPU memory of the index tables of the size and in all, with the peak of the size's first decode, in GB and bytes per pixel; the time of that first decode, everything included, and of later decodes (minimum of 3 rounds), in ms, and megapixels per second of the later ones; untuned streams of a DIV2K photo resized with Lanczos, lanes of 2,048 values (16\x2, 64\x2) and 8,192 (512\x4). Every decode is exact. The 512\x4's decodes of 33 megapixels heat the GPU until our machine's temperature guard pauses them, so their time is an upper bound (on an A100: 2.56\,s).}
\label{S:tab:large}
\end{table}

\subsection{Index tables}
\label{S:sec:tables}
The tables that depend only on the image size, such as the positions of the values of each step and their neighbours, are built once per size and kept, and are not in the reported times. The decoder of the main paper's runs builds them on the host, in 26\,ms for Kodak's sizes and 122--123\,ms for a 2K photo with the RTX 3090's host CPU (\cref{S:tab:cold}), and its parser also builds the CPU decoder's tables (\cref{S:sec:memory}), 0.4\,s for a Kodak photo and 4.3\,s for a 2K photo. Every entry is a function of the value's index alone, so one CUDA kernel, one thread per value, builds the same tables, which we check entry for entry on 13 sizes from $65\times63$ to $7680\times4320$: in 1.9--3.4\,ms for Kodak, 7.2\,ms for 2K and 69\,ms for 33 megapixels (1.8\,s on the host). With it and the light parse, the first decode of a size, everything included, takes about 1\,ms longer than a later one for a Kodak photo with our 16\x2 (7.1--7.2 against 6.0--6.1\,ms) and 5.7\,ms longer for a 2K photo (19.6 against 13.9\,ms), and 0.7--5.9\,ms longer with our 64\x2 and 512\x4 (\cref{S:tab:cold}); a process whose memory allocator has released its cache also allocates the tables and buffers again, 3.3--3.5\,ms more for a Kodak photo and 19\,ms for a 2K photo. A set of one size pays this once, and a set of many sizes once per size: the 11 CLIC.p photos of the main paper's Tab.~3 have 8 sizes. The dense mode has no index tables, only staging buffers per image layout: on the same GPU, its first decode of a size, buffers included and without a CUDA graph, takes 1.1\,ms longer than a later one for a Kodak photo (median over the 24 photos: 2.0 against 0.61\,ms) and 3.2\,ms longer for a CLIC.p or DIV2K/4 photo (4.8--5.0 against 1.7--1.8\,ms); recording the graph that later decodes of the size replay adds 0.8--1.7\,ms more.

\begin{table}[t]
\centering
\footnotesize
\setlength{\tabcolsep}{2.6pt}
\input{supp/tab_cold}
\caption{\textbf{Index tables and the first decode of a size} on one RTX 3090, milliseconds: the size's tables built on the host (our decoder's C++ code) and on the GPU (one kernel, the same tables), and, per network, the first decode of the size with the GPU's tables and the light parse, from emptied caches of tables and everything included, and a later decode (minimum of 5, each after a short GPU warm-up). Untuned streams; every decode is exact. Kodak: kodim01; 2K: DIV2K 0801.}
\label{S:tab:cold}
\end{table}

\subsection{A corpus of many sizes}
\label{S:sec:coco}
To see what keeping tables per size costs when the size changes from one image to the next, we code the 5,000 photos of COCO val2017, JPEG photos decoded once to RGB, in 603 sizes (350 of them used once; 0.27 megapixels on average), with untuned streams, and decode them one at a time in the order of their file names on the RTX 3090 (\cref{S:tab:coco}): with the tables of every size ready (each photo decoded once and checked, then three times after a short GPU warm-up, the fastest counted), with those of the 16 most recently used sizes kept, which a new size replaces 2,256 times, and with tables built for every photo. Keeping 16 sizes, in 0.9\,GB, our 16\x2 and 64\x2 decode a photo in 7.2 and 12.4\,ms, at most 1.0 and 1.4\,ms more than with every size ready; building the tables for every photo adds 0.5 and 0.2\,ms more. The dense mode needs nothing kept: decoding each photo without a CUDA graph, with staging buffers made for its stream, takes 0.80\,ms, less than replaying the graphs recorded for each size (1.4\,ms with every size ready), whose buffers must hold any stream of their size. On every fifth photo (1,000, encoded on the A100), our 64\x2 makes files 0.9\% larger than JPEG XL e9 untuned (the best of four colour transforms and eight orientations; 2.6\% on Kodak) and, tuned to each photo, 1.0\% smaller (95\% bootstrap interval 0.9--1.2\%), smaller on 672 of the 1,000.

\begin{table}[t]
\centering
\footnotesize
\setlength{\tabcolsep}{3pt}
\input{supp/tab_coco}
\caption{\textbf{A corpus of many sizes}: the 5,000 photos of COCO val2017 (603 sizes), untuned streams, decoded one at a time in the order of their file names on one RTX 3090, from the stream in host memory to the image in GPU memory: time per photo with the tables (dense: staging buffers, and the CUDA graph it replays) of every size ready, of the 16 most recently used sizes (2,256 builds), or of none. Every decode is exact. JPEG XL e9 and WebP (lossless) for reference, on the CPU.}
\label{S:tab:coco}
\end{table}

\subsection{Host and GPU work}
\label{S:sec:host}
HTJ2K's decoder, nvJPEG2000, and our dense decoder both parse a stream on the host and decode it on the GPU. To see how much of a single-photo decode each spends on either side, we trace every kernel, copy and memset of the process with CUPTI while each decodes one photo at a time, on the RTX 3090 of our CPU machine, whose i9-14900K is a faster host than our server's EPYC 7702 (\cref{S:tab:host}). Both keep the GPU busy for most of a decode: for 0.60 of HTJ2K's 0.73\,ms per Kodak photo and for 0.43 of the dense mode's 0.54\,ms, and for 73--75\% and 81\% of their times on the large photos (busy times from the traced runs, decode times from untraced ones: tracing slows the host's calls, not the GPU's work). The host's share, the time in which the GPU waits, is thus 0.13--0.37\,ms per photo for HTJ2K and 0.11--0.26\,ms for the dense mode. Tracing adds 0.8--1.0\,ms per photo to HTJ2K but only 0.04--0.06\,ms to the dense mode, which replays one CUDA graph: nvJPEG2000 makes many more calls to the GPU driver. On this machine the dense mode is 1.1--1.4\x\ faster than HTJ2K in two sessions, with the files of the main paper; on our server, with a slower host and a faster GPU, it is 1.2--1.6\x\ faster (main paper, Tab.~3), and a profile on CLIC.p there shows the GPU working for about half of the dense mode's 0.83\,ms per photo (0.39\,ms).

\begin{table}[t]
\centering
\footnotesize
\setlength{\tabcolsep}{2.6pt}
\input{supp/tab_host}
\caption{\textbf{Host and GPU work} of single-photo decodes on one RTX 3090 with an i9-14900K host, the files of the main paper: milliseconds per photo; GPU: the time in which the GPU is busy, the union of every kernel, copy and memset of the process inside the decode, from a CUPTI trace, and its share of the decode time; traced: the decode time under tracing. Every decode is exact.}
\label{S:tab:host}
\end{table}

\section{Every codec, one image at a time}
\label{S:sec:everycodec}

\begin{table*}[t]
\centering
\footnotesize
\setlength{\tabcolsep}{5pt}
\begin{tabular}{@{}llrrrrrr@{}}
\toprule
& & \multicolumn{2}{c}{Kodak} & \multicolumn{2}{c}{CLIC.p / 4} & \multicolumn{2}{c}{DIV2K / 4} \\
\cmidrule(lr){3-4}\cmidrule(lr){5-6}\cmidrule(l){7-8}
Codec & on & bpsp & ms & bpsp & ms & bpsp & ms \\
\midrule
\fmt fast & GPU & 3.981 & 0.23 / 0.25 & 3.870 & 0.71 / 0.74 & 4.145 & 0.87 / 0.93 \\
\fmt dense & GPU & 3.187 & 0.45 / 0.46 & 2.671 & 0.83 / 0.84 & 3.071 & 1.02 / 1.06 \\
\fmt 16\x2 & GPU & 2.960 & 4.54 / 4.59 & 2.447 & 7.91 / 8.06 & 2.782 & 8.92 / 9.69 \\
\fmt 32\x2 & GPU & 2.902 & 4.92 / 4.94 & 2.408 & 8.91 / 8.94 & 2.731 & 10.0 / 10.9 \\
\fmt 64\x2 & GPU & 2.818 & 7.49 / 7.51 & 2.364 & 16.9 / 17.0 & 2.679 & 19.1 / 20.0 \\
\fmt 128\x2 & GPU & 2.779 & 33.1 / 33.3 & -- & -- & -- & -- \\
\fmt 512\x4 & GPU & 2.589 & 53.5 / 53.6 & 2.239 & 233 / 234 & 2.545 & 270 / 272 \\
\fmt 768\x4 & GPU & 2.589 & 76.5 / 76.7 & 2.236 & 372 / 374 & 2.540 & 432 / 435 \\
TIFF-LZW (nvTIFF) & GPU & 5.078 & 0.19 / 0.21 & 4.226 & 0.52 / 0.62 & 5.204 & 0.77 / 0.81 \\
nvCOMP Bitcomp & GPU & 6.904 & 0.23 / 0.27 & 6.352 & 0.45 / 0.53 & 6.696 & 0.50 / 0.55 \\
HTJ2K (nvJPEG2000) & GPU & 3.510 & 0.55 / 0.57 & 2.960 & 1.30 / 1.42 & 3.354 & 1.54 / 1.60 \\
JPEG 2000 (nvJPEG2000) & GPU & 3.191 & 6.22 / 6.24 & 2.760 & 7.54 / 7.61 & 3.132 & 8.99 / 9.02 \\
DLPR, run by us$^\dagger$ & GPU & 2.857 & 5373 & 2.241 & 14280 & 2.522 & 16189 \\
\midrule
\fmt fast & CPU & 3.981 & 0.69 / 0.81 & 3.870 & 4.32 / 4.68 & 4.145 & 4.98 / 5.49 \\
\fmt 32\x2, 8-bit & CPU & 2.906 & 11.8 / 12.2 & 2.403 & 65.2 / 68.0 & 2.747 & 91.3 / 93.5 \\
\fmt 64\x2, 8-bit & CPU & 2.825 & 16.7 / 16.9 & 2.359 & 93.6 / 97.2 & 2.677 & 117 / 119 \\
QOI & CPU & 4.663 & 1.50 / 1.58 & 4.259 & 9.79 / 10.2 & 4.883 & 11.0 / 11.3 \\
PNG & CPU & 4.457 & 5.07 / 5.20 & 3.649 & 26.5 / 27.8 & 4.348 & 36.7 / 37.3 \\
WebP & CPU & 3.166 & 4.63 / 4.78 & 2.751 & 27.8 / 29.1 & 3.154 & 35.3 / 35.8 \\
JPEG-LS (CharLS) & CPU & 3.213 & 11.1 / 11.1 & 2.749 & 58.2 / 60.6 & 3.104 & 78.1 / 79.0 \\
JPEG 2000 (OpenJPEG) & CPU & 3.191 & 8.85 / 9.00 & 2.760 & 27.6 / 29.2 & 3.132 & 33.9 / 34.9 \\
AVIF & CPU & 3.795 & 26.0 / 26.3 & 3.369 & 142 / 146 & 3.838 & 192 / 194 \\
JPEG XL e1 & CPU & 3.640 & 1.81 / 1.91 & 3.084 & 5.16 / 5.59 & 3.462 & 5.81 / 6.12 \\
JPEG XL e3 & CPU & 3.146 & 4.48 / 4.61 & 2.627 & 12.3 / 12.8 & 2.929 & 14.2 / 14.6 \\
JPEG XL e5 & CPU & 3.059 & 9.28 / 9.49 & 2.569 & 18.9 / 20.0 & 2.875 & 22.8 / 23.5 \\
JPEG XL e7 & CPU & 2.902 & 11.7 / 12.0 & 2.485 & 23.7 / 24.6 & 2.816 & 28.0 / 28.7 \\
JPEG XL e9 & CPU & 2.885 & 10.7 / 11.0 & 2.457 & 25.1 / 25.9 & 2.785 & 29.8 / 30.4 \\
JPEG XL e10$^\ddagger$ & CPU & 2.865 & 11.3 / 11.6 & -- & -- & -- & -- \\
\bottomrule
\end{tabular}
\caption{\textbf{Every codec we ran, one image at a time}: bits per subpixel and decode time per image (minimum / median over 7 rounds, each averaged over the images), from compressed bytes to pixels, on one RTX 4090 or on the i9-14900K at each codec's best number of threads (Kodak JPEG-LS: the re-measurement of Fig.~\ref{S:fig:large}c). $^\dagger$: mean of one decode per image. $^\ddagger$: Kodak only, in a later session (libjxl 0.11.2, as for every JPEG XL row), in which e9 took 11.1 / 11.3\,ms; through imagecodecs, effort 11 writes the same files as effort 10 (libjxl runs it only with expert options), so we report effort 10. Settings: HTJ2K by OpenJPH and JPEG 2000 by OpenJPEG, both through imagecodecs with their defaults for reversible coding (the 5/3 wavelet with 5 decomposition levels, code-blocks of $64\times64$, the reversible colour transform); JPEG XL decoded with libjxl's default options (no faster-decoding mode, which its encoder would have to write into the file). With medians instead of minima, every codec that is larger and slower than an \fmt mode stays so; nvCOMP on Kodak and JPEG XL e3 (against our 16\x2) become so as well. Learned codecs with published times are in the main paper (Tab.~4).}
\label{S:tab:lat_all}
\end{table*}

Table~\ref{S:tab:lat_all} lists the sizes and the single-image decode times behind the frontiers of the main paper (Fig.~4) and of
Fig.~\ref{S:fig:large}, and Table~\ref{S:tab:tput_all} the throughput of every codec at full load.

\section{Throughput of every codec}
\label{S:sec:throughput}

\cref{S:tab:tput_all} gives the throughput of every codec we ran at full load. Of nvCOMP we ran eight compressors, Deflate, GDeflate, ANS and Bitcomp on raw pixels and after PNG's Sub filter, each with chunks of 16 or 64\,KB; \cref{S:tab:nvcomp} lists those on nvCOMP's own size--speed frontier. Our dense mode at full load (3.150, 2.635 and 3.034 \bpsp; 6,716, 7,886 and 8,019\,MP/s) is smaller and faster than every one of them.
\begin{table}[t]
\centering
\footnotesize
\setlength{\tabcolsep}{4pt}
\input{supp/tab_nvcomp}
\caption{\textbf{nvCOMP at full load} on one RTX 4090: the variants on its own size--speed frontier (algorithm,
filter, chunk size), bits per subpixel and megapixels per second, best number of host threads and batch size.}
\label{S:tab:nvcomp}
\end{table}

\begin{table}[h]
\centering
\footnotesize
\setlength{\tabcolsep}{3pt}
\begin{tabular}{@{}llrrr@{}}
\toprule
Codec & device & Kodak & CLIC.p / 4 & DIV2K / 4 \\
\midrule
\fmt fast & GPU & 8811 & 11792 & 9116 \\
\fmt dense ($T=2048$) & GPU & 6716 & 7886 & 8019 \\
HTJ2K (nvJPEG2000) & GPU & 4533 & 7097 & 7013 \\
TIFF-LZW (nvTIFF) & GPU & 4558 & 6105 & 5587 \\
nvCOMP, fastest & GPU & 2205 & 3364 & 3139 \\
JPEG 2000 (nvJPEG2000) & GPU & 410 & 1122 & 1056 \\
\fmt 16\x2 & GPU & 397 & 490 & 490 \\
\fmt 32\x2 & GPU & 401 & 402 & 410 \\
\fmt 64\x2 & GPU & 161 & 185 & 188 \\
\fmt 512\x4 & GPU & 7 & 10 & 10 \\
\fmt 768\x4 & GPU & 5 & 6 & 6 \\
QOI & CPU & 3077 & 3195 & 3512 \\
WebP & CPU & 1664 & 1771 & 1690 \\
PNG & CPU & 1572 & 1813 & 1596 \\
JPEG XL e1 & CPU & 886 & 977 & 977 \\
\fmt fast & CPU & 838 & 748 & 762 \\
JPEG-LS (CharLS) & CPU & 687 & 782 & 700 \\
AVIF & CPU & 305 & 332 & 294 \\
JPEG XL e3 & CPU & 298 & 310 & 301 \\
JPEG 2000 (OpenJPEG) & CPU & 189 & 206 & 169 \\
JPEG XL e5 & CPU & 150 & 189 & 175 \\
JPEG XL e9 & CPU & 135 & 142 & 135 \\
JPEG XL e7 & CPU & 123 & 149 & 143 \\
\bottomrule
\end{tabular}
\caption{\textbf{Throughput at full load} of every codec we ran, in megapixels per second (best number of host threads and batch size on one RTX 4090; all 32 threads of the i9-14900K). nvCOMP: Bitcomp on Kodak and DIV2K, GDeflate on CLIC.p. Our 8-bit 32\x2 and 64\x2 CPU decoders reach 38 and 26\,MP/s. At full load our dense mode, with lanes of 2,048 values (3.150 \bpsp on Kodak), is 0.1\% larger than JPEG XL e3, but our 16\x2 is smaller and faster than JPEG XL e3 on every set. Rows by Kodak throughput.}
\label{S:tab:tput_all}
\end{table}

\section{Potential directions for further research}
\label{S:sec:future}

\paragraph{From images to video.}
A video adds a time axis to the tree. In a frame predicted from the previous one, every level can also read the decoded pixels of the previous frame at and around the same position, so static and slowly moving regions cost almost nothing, while the number of steps per frame stays the same. Our measurements on large photos give an idea of the speed: one image at a time, the dense mode decodes a 2.4-megapixel photo, a little larger than a 1080p frame, in 0.83\,ms, and the 64\x2 network in 17\,ms, about 60 frames per second; at full load, the dense mode decodes 7.9 gigapixels per second, as many pixels as 950 frames of 4K video per second. Lossless video is stored today with intra-frame codecs such as FFV1~\cite{ffv1}, decoded on the CPU. A tree that also predicts across time could bring learned compression to video masters, medical and scientific recordings and lossless streaming. Because the coarse levels come first, a stream whose lanes end with each level would also let a client show a frame before it has fully arrived, and drop its finest levels when the bandwidth falls short.

\paragraph{Tuning in one pass.} Our encoder tunes a small update of its networks to each photo by gradient descent, the largest single stage of the 512\x4's encoding (\cref{S:sec:fastenc}). A network trained to predict this update from the photo in one pass, a hypernetwork, would replace the iterations, as amortised inference replaces per-example optimisation elsewhere; the encoder would still keep the predicted update of a colour channel only where its exact code length shortens the file, as it does now, and could refine it with a few iterations.

\paragraph{Beyond compression.}
To decode a pixel, a neural mode computes features of its decoded neighbourhood, so decoding also yields image features as a by-product. On CIFAR-10, whose $32\times32$ images make a tree of one root pixel and five levels, a linear classifier on the activations of our 512\x4 predictor, with its bits and residuals, averaged per tree level and image quadrant, reaches 75.7\% accuracy, against 62.3\% for the same network with random weights and 41.8\% for raw pixels; our 64\x2 reaches 68.6\% (with random weights, 60.6\%). Training to compress thus adds 8--13 points over random features of the same size, in line with evidence that models trained to predict pixels learn useful representations~\cite{igpt}. Recognising images from a partly decoded stream, perhaps from its coarse levels alone, is a potential direction for further research.

%% file: supp/tab_lanes.tex
\begin{tabular}{@{}rrrrrrrrrrrr@{}}
\toprule
& & & \multicolumn{3}{c}{16\x2} & \multicolumn{3}{c}{64\x2} & \multicolumn{3}{c}{512\x4} \\
\cmidrule(lr){4-6}\cmidrule(lr){7-9}\cmidrule(l){10-12}
$T$ & $G$ & per lane & bpsp & dec. & ent. & bpsp & dec. & ent. & bpsp & dec. & ent. \\
\midrule
512 & 72 & 512 & 2.997 & 5.4 & 4.0 & 2.855 & 11.5 & 4.2 & 2.634 & 80.3 & 4.9 \\
2,048 & 18 & 2,048 & 2.961 & 6.5 & 5.1 & 2.819 & 12.6 & 5.4 & 2.597 & 81.3 & 5.8 \\
8,192 & 5 & 7,371 & 2.952 & 10.1 & 8.7 & 2.810 & 16.1 & 8.8 & 2.588 & 84.7 & 9.3 \\
32,768 & 2 & 18,428 & 2.950 & 17.6 & 16.2 & 2.808 & 23.3 & 16.0 & 2.586 & 91.7 & 16.5 \\
262,144 & 1 & 36,855 & 2.949 & 29.6 & 28.3 & 2.807 & 35.0 & 27.8 & 2.585 & 102.9 & 28.0 \\
\bottomrule
\end{tabular}

%% file: supp/tab_order.tex
\begin{tabular}{@{}lrrrrrr@{}}
\toprule
 & & \multicolumn{2}{c}{ideal, \bpsp} & file, & \multicolumn{2}{c}{GPU part, ms} \\
\cmidrule(lr){3-4}\cmidrule(lr){6-7}
order & steps & generic & tuned & \bpsp & codec & graph \\
\midrule
\multicolumn{7}{@{}l}{\emph{Final networks: 60,000 + 120,000 training steps (our 16\x2: the first 60,000 on the RTX 3090, the rest on the RTX 4090 server)}} \\
tree (ours) & 72 & 3.084 & 2.949 & 2.960 & 6.5 & 5.2 \\
2:1 wavefront & 5{,}367--6{,}135 & 3.007 ($-$2.5\%) & 2.900 ($-$1.7\%) & 2.913 ($-$1.6\%) & 258.4 & 235.2 \\
raster & 1{,}179{,}360 & 3.015 ($-$2.2\%) & 2.929 ($-$0.7\%) & -- & -- & -- \\
\midrule
\multicolumn{7}{@{}l}{\emph{First training cycle: 60,000 steps, all three trained on the RTX 3090}} \\
tree & 72 & 3.087 & 2.955 & -- & -- & -- \\
2:1 wavefront & 5{,}367--6{,}135 & 3.015 ($-$2.3\%) & 2.905 ($-$1.7\%) & -- & -- & -- \\
raster & 1{,}179{,}360 & 3.025 ($-$2.0\%) & 2.922 ($-$1.1\%) & -- & -- & -- \\
\midrule
\multicolumn{7}{@{}l}{\emph{64\x2, first training cycle: 60,000 steps on the RTX 3090 (tree: the first cycle of our 64\x2)}} \\
tree & 72 & 2.985 & 2.837 & -- & -- & -- \\
2:1 wavefront & 5{,}367--6{,}135 & 2.887 ($-$3.3\%) & 2.758 ($-$2.8\%) & -- & -- & -- \\
\midrule
\multicolumn{7}{@{}l}{\emph{64\x2, final networks: 60,000 + 120,000 steps (tree: our 64\x2; wavefront: both cycles on the RTX 3090)}} \\
tree (ours) & 72 & 2.962 & 2.808 & -- & -- & -- \\
2:1 wavefront & 5{,}367--6{,}135 & 2.874 ($-$3.0\%) & 2.740 ($-$2.4\%) & -- & -- & -- \\
\midrule
\multicolumn{7}{@{}l}{\emph{128\x2, one training cycle: 60,000 steps (tree: our 128\x2, on the RTX 4090 server; wavefront: on the RTX 3090)}} \\
tree (ours) & 72 & 2.926 & 2.750 & -- & -- & -- \\
2:1 wavefront & 5{,}367--6{,}135 & 2.847 ($-$2.7\%) & 2.720 ($-$1.1\%) & -- & -- & -- \\
\midrule
\multicolumn{7}{@{}l}{\emph{512\x4, one matched short recipe: 30,000 steps from scratch, both on the RTX 3090}} \\
tree & 72 & 2.852 & 2.713 & -- & -- & -- \\
2:1 wavefront & 5{,}367--6{,}135 & 2.803 ($-$1.7\%) & 2.726 ($+$0.5\%) & -- & -- & -- \\
\midrule
\multicolumn{7}{@{}l}{\emph{16\x2, final networks, one session: the ideal code lengths of every row computed anew; the wavefront networks above, inside tiles without retraining}} \\
tree (ours) & 72 & 3.084 & 2.949 & 2.961 & 6.5 & 5.2 \\
2:1 wavefront, 64-px tiles & 567 & 3.049 ($-$1.1\%) & 2.941 ($-$0.3\%) & 2.951 ($-$0.3\%) & 29.6 & 26.3 \\
2:1 wavefront, 128-px tiles & 1{,}143 & 3.025 ($-$1.9\%) & 2.920 ($-$1.0\%) & 2.930 ($-$1.0\%) & 56.4 & 50.8 \\
2:1 wavefront, 256-px tiles & 2{,}295 & 3.014 ($-$2.3\%) & 2.907 ($-$1.4\%) & 2.918 ($-$1.4\%) & 109.5 & 99.4 \\
2:1 wavefront, whole image & 5{,}367--6{,}135 & 3.007 ($-$2.5\%) & 2.900 ($-$1.7\%) & -- & -- & -- \\
\midrule
\multicolumn{7}{@{}l}{\emph{64\x2, final networks, one session: the ideal code lengths of every row computed anew; the wavefront networks above, inside tiles without retraining}} \\
tree (ours) & 72 & 2.962 & 2.803 & -- & -- & -- \\
2:1 wavefront, 64-px tiles & 567 & 2.917 ($-$1.5\%) & 2.779 ($-$0.8\%) & -- & -- & -- \\
2:1 wavefront, 128-px tiles & 1{,}143 & 2.893 ($-$2.3\%) & 2.758 ($-$1.6\%) & -- & -- & -- \\
2:1 wavefront, 256-px tiles & 2{,}295 & 2.881 ($-$2.7\%) & 2.746 ($-$2.0\%) & -- & -- & -- \\
2:1 wavefront, whole image & 5{,}367--6{,}135 & 2.874 ($-$3.0\%) & 2.739 ($-$2.3\%) & -- & -- & -- \\
\bottomrule
\end{tabular}

%% file: supp/tab_root.tex
\begin{tabular}{@{}rrrrrr@{}}
\toprule
$S$ & levels & steps & root & generic & tuned \\
\midrule
32 & 5 & 60 & 1/1{,}024 & 3.092 ($+$0.2\%) & 2.968 ($+$0.5\%) \\
64 & 6 & 72 & 1/4{,}096 & 3.086 & 2.954 \\
128 & 7 & 84 & 1/16{,}384 & 3.087 ($+$0.0\%) & 2.949 ($-$0.2\%) \\
\bottomrule
\end{tabular}

%% file: supp/tab_ablation.tex
\begin{tabular}{@{}lrrrrrr@{}}
\toprule
Encoder & \multicolumn{2}{c}{16\x2} & \multicolumn{2}{c}{64\x2} & \multicolumn{2}{c}{512\x4} \\
\cmidrule(lr){2-3}\cmidrule(lr){4-5}\cmidrule(l){6-7}
 & bpsp & $\Delta$ & bpsp & $\Delta$ & bpsp & $\Delta$ \\
\midrule
first-ranked transform & 3.096 &  & 2.970 &  & 2.783 &  \\
+ best of top four & 3.088 & $-0.3$\% & 2.967 & $-0.1$\% & 2.777 & $-0.2$\% \\
+ best of 8 orientations & 3.075 & $-0.4$\% & 2.960 & $-0.2$\% & 2.774 & $-0.1$\% \\
+ per-photo tuning & 2.961 & $-3.7$\% & 2.819 & $-4.8$\% & 2.588 & $-6.7$\% \\
\bottomrule
\end{tabular}

%% file: supp/tab_untuned.tex
\begin{tabular}{@{}lrrrrrrrrr@{}}
\toprule
 & \multicolumn{3}{c}{Kodak} & \multicolumn{3}{c}{CLIC.p / 4} & \multicolumn{3}{c}{DIV2K / 4} \\
\cmidrule(lr){2-4}\cmidrule(lr){5-7}\cmidrule(l){8-10}
Network & untuned & tuned & gain & untuned & tuned & gain & untuned & tuned & gain \\
\midrule
16\x2 & 3.075 & 2.960 & 3.7\% & 2.492 & 2.447 & 1.8\% & 2.828 & 2.782 & 1.6\% \\
32\x2 & 3.031 & 2.902 & 4.3\% & 2.446 & 2.408 & 1.6\% & 2.773 & 2.731 & 1.5\% \\
64\x2 & 2.960 & 2.818 & 4.8\% & 2.397 & 2.364 & 1.4\% & 2.717 & 2.679 & 1.4\% \\
128\x2 & 2.927 & 2.779 & 5.1\% & -- & -- & -- & -- & -- & -- \\
512\x4 & 2.774 & 2.589 & 6.7\% & 2.263 & 2.239 & 1.1\% & 2.576 & 2.545 & 1.2\% \\
768\x4 & 2.769 & 2.589 & 6.5\% & 2.259 & 2.236 & 1.0\% & 2.570 & 2.540 & 1.2\% \\
\bottomrule
\end{tabular}

%% file: supp/tab_encfast.tex
\begin{tabular}{@{}llrrrrrrrrrr@{}}
\toprule
 & & \multicolumn{2}{c}{Kodak, untuned} & \multicolumn{4}{c}{Kodak, tuned} & \multicolumn{4}{c}{2K photos, tuned} \\
\cmidrule(lr){3-4}\cmidrule(lr){5-8}\cmidrule(l){9-12}
GPU & Network & paper & same & paper & same & fast & pool & paper & same & fast & pool \\
\midrule
RTX 3090 & 16\x2 & 1.3 & 0.06 & 4.7 & 3.7 & 2.3 ($-$0.02\%) & 0.85 ($+$0.03\%) & 30.2 & 10.7 & 3.2 (0\%) & 2.0 ($+$0.01\%) \\
 & 64\x2 & 1.4 & 0.04 & 5.2 & 4.1 & 2.9 ($+$0.05\%) & 1.1 ($+$0.16\%) & 31.0 & 11.4 & 4.1 (0\%) & 2.6 ($+$0.03\%) \\
 & 512\x4 & -- & -- & -- & -- & -- & 5.0 ($+$0.14\%) & -- & -- & -- & 13.9 ($+$0.05\%) \\
\addlinespace[2pt]
A100 & 16\x2 & -- & -- & -- & -- & 2.1 ($-$0.01\%) & 1.1 ($+$0.01\%) & 33.5 & -- & 2.6 (0\%) & 2.6 ($+$0.03\%) \\
 & 64\x2 & -- & -- & -- & -- & 2.2 ($+$0.01\%) & 1.0 ($+$0.12\%) & 33.8 & -- & 3.3 (0\%) & 2.7 ($+$0.03\%) \\
 & 512\x4 & -- & -- & -- & -- & 8.9 ($+$0.03\%) & 3.5 ($+$0.15\%) & 41.4 & -- & 13.6 (0\%) & 7.5 ($+$0.08\%) \\
 & 768\x4 & -- & -- & -- & -- & -- & 4.6 ($+$0.11\%) & -- & -- & -- & -- \\
\bottomrule
\end{tabular}

%% file: supp/tab_encbreak.tex
\begin{tabular}{@{}llrrrrrrrr@{}}
\toprule
Set & Network & rank & tables & top 4 & orient. & tuning & quant. & coding & total \\
\midrule
CLIC.p & 16\x2 & 10.6 & 8.0 & 1.4 & 2.3 & 2.5 & 3.9 & 0.3 & 29.0 \\
 & 64\x2 & 10.5 & 8.1 & 1.3 & 2.3 & 2.7 & 4.0 & 0.3 & 29.1 \\
 & 512\x4 & 10.6 & 8.0 & 2.5 & 4.4 & 12.0 & 5.8 & 0.8 & 44.0 \\
\addlinespace[2pt]
DIV2K & 16\x2 & 11.1 & 8.7 & 1.5 & 2.3 & 2.6 & 4.1 & 0.4 & 30.6 \\
 & 64\x2 & 11.1 & 8.6 & 1.3 & 2.4 & 2.7 & 4.2 & 0.3 & 30.6 \\
 & 512\x4 & 11.1 & 8.6 & 2.6 & 4.6 & 12.0 & 6.1 & 0.8 & 45.9 \\
\bottomrule
\end{tabular}

%% file: supp/tab_thumbtime.tex
\begin{tabular}{@{}lrrrrrr@{}}
\toprule
 & \multicolumn{2}{c}{64$^2$} & \multicolumn{2}{c}{128$^2$} & \multicolumn{2}{c}{256$^2$} \\
\cmidrule(lr){2-3}\cmidrule(lr){4-5}\cmidrule(l){6-7}
Codec & bpsp & ms & bpsp & ms & bpsp & ms \\
\midrule
\multicolumn{7}{@{}l}{\emph{\fmt, untuned, on the RTX 3090}} \\
16\x2 & 3.480 & 3.2 & 3.368 & 4.1 & 3.294 & 5.5 \\
64\x2 & 3.350 & 5.7 & 3.241 & 6.8 & 3.166 & 8.1 \\
512\x4 & 3.160 & 23.9 & 3.063 & 24.8 & 2.981 & 28.6 \\
\multicolumn{7}{@{}l}{\emph{One thread of the i9-14900K}} \\
PNG & 5.072 & 0.06 & 4.954 & 0.24 & 4.802 & 0.92 \\
WebP & 3.688 & 0.07 & 3.546 & 0.22 & 3.392 & 0.82 \\
JPEG XL e3 & 3.642 & 0.30 & 3.479 & 1.1 & 3.365 & 3.7 \\
JPEG XL e9 & 3.454 & 0.53 & 3.314 & 2.4 & 3.170 & 10.7 \\
\bottomrule
\end{tabular}

%% file: supp/tab_thumbs.tex
\begin{tabular}{@{}lrrrrr@{}}
\toprule
Network & crop & untuned & tuned & kept & update \\
\midrule
16\x2 & 64$^2$ & 3.452 & 3.395 ($-$1.6\%) & 89\% & 1.7\% \\
 & 128$^2$ & 3.351 & 3.257 ($-$2.8\%) & 99\% & 1.2\% \\
 & 256$^2$ & 3.276 & 3.164 ($-$3.4\%) & 100\% & 0.7\% \\
\addlinespace[2pt]
64\x2 & 64$^2$ & 3.330 & 3.297 ($-$1.0\%) & 72\% & 2.1\% \\
 & 128$^2$ & 3.230 & 3.149 ($-$2.5\%) & 100\% & 1.7\% \\
 & 256$^2$ & 3.155 & 3.036 ($-$3.8\%) & 100\% & 1.1\% \\
\addlinespace[2pt]
512\x4 & 64$^2$ & 3.150 & 3.122 ($-$0.9\%) & 56\% & 1.8\% \\
 & 128$^2$ & 3.052 & 2.982 ($-$2.3\%) & 85\% & 1.9\% \\
 & 256$^2$ & 2.981 & 2.863 ($-$4.0\%) & 100\% & 1.7\% \\
\bottomrule
\end{tabular}

%% file: supp/tab_encode.tex
\begin{tabular}{@{}lrrrrrr@{}}
\toprule
Tuning & \multicolumn{2}{c}{16\x2} & \multicolumn{2}{c}{64\x2} & \multicolumn{2}{c}{512\x4} \\
\cmidrule(lr){2-3}\cmidrule(lr){4-5}\cmidrule(l){6-7}
 & bpsp & s & bpsp & s & bpsp & s \\
\midrule
200 iterations & 2.961 & 4.5 & 2.819 & 5.1 & 2.588 & 14.9 \\
25 iterations & 2.995 & 2.6 & 2.868 & 2.7 & 2.676 & 4.4 \\
\bottomrule
\end{tabular}

%% file: supp/tab_port.tex
\begin{tabular}{@{}lrrrrcc@{}}
\toprule
 & & \multicolumn{3}{c}{A100, ms} & \multicolumn{2}{c}{same streams} \\
\cmidrule(lr){3-5}\cmidrule(l){6-7}
Network & size & std. & port. & ratio & std. & port. \\
\midrule
16\x2 & $-$0.0001\% & 7.1 & 7.4 & 1.04 & 24/24 & 24/24 \\
32\x2 & $-$0.0001\% & 8.0 & 8.2 & 1.03 & 24/24 & 24/24 \\
64\x2 & $-$0.0001\% & 12.1 & 12.7 & 1.05 & 24/24 & 24/24 \\
128\x2 & $+$0.0045\% & 28.1 & 26.4 & 0.94 & 0/24 & 24/24 \\
512\x4 & $-$0.0017\% & 55.1 & 167 & 3.03 & 0/24 & 24/24 \\
768\x4 & $+$0.0023\% & 70.6 & 311 & 4.41 & 0/24 & 24/24 \\
\bottomrule
\end{tabular}

%% file: supp/tab_port8.tex
\begin{tabular}{@{}llrrrrrc@{}}
\toprule
 & & & \multicolumn{2}{c}{A100} & \multicolumn{2}{c}{RTX 3090} & \\
\cmidrule(lr){4-5}\cmidrule(lr){6-7}
Network & kernels & size & ms & ratio & ms & ratio & same \\
\midrule
512\x4 & standard & -- & 81.0 & 1.00 & 84.6 & 1.00 & 0/24 \\
 & fp32 & $-$0.002\% & 168 & 2.07 & 143 & 1.68 & 24/24 \\
 & int8 & $+$0.25\% & 89.0 & 1.10 & 85.3 & 1.01 & 24/24 \\
 & int8, asym. & $+$0.16\% & 91.2 & 1.12 & 87.3 & 1.03 & 24/24 \\
\midrule
768\x4 & standard & -- & 85.9 & 1.00 & 130 & 1.00 & 0/24 \\
 & fp32 & $+$0.002\% & 312 & 3.64 & 267 & 2.05 & 24/24 \\
 & int8 & $+$0.20\% & 127 & 1.47 & 125 & 0.96 & 24/24 \\
 & int8, asym. & $+$0.11\% & 129 & 1.50 & 127 & 0.98 & 24/24 \\
\midrule
\multicolumn{8}{@{}l}{\emph{Tuned, encoder tuning through the integer layers}} \\
512\x4 & standard & -- & & & 89.4 & 1.00 & \\
 & int8, asym. & $+$0.25\% & & & 91.7 & 1.03 & \\
768\x4 & standard & -- & & & 134.8 & 1.00 & \\
 & int8, asym. & $+$0.21\% & & & 131.8 & 0.98 & \\
\bottomrule
\end{tabular}

%% file: supp/tab_xgpu.tex
\begin{tabular}{@{}lrcr@{}}
\toprule
Mode & bpsp & exact & ms \\
\midrule
fast & 3.981 & yes & 0.19 \\
dense & 3.187 & yes & 0.54 \\
16\x2 & 2.960 & yes & 6.33 \\
32\x2 & 2.902 & yes & 7.15 \\
64\x2 & 2.818 & yes & 12.5 \\
128\x2 & 2.779 & no & -- \\
512\x4 & 2.589 & no & -- \\
768\x4 & 2.589 & no & -- \\
\bottomrule
\end{tabular}

%% file: supp/tab_means.tex
\begin{tabular}{@{}lrrrrrr@{}}
\toprule
 & \multicolumn{3}{c}{CLIC.p (41)} & \multicolumn{3}{c}{DIV2K (100)} \\
\cmidrule(lr){2-4}\cmidrule(l){5-7}
Codec & all bits & per photo & published & all bits & per photo & published \\
\midrule
\fmt 16\x2 & 2.602 & 2.671 &  & 2.775 & 2.783 &  \\
\fmt 64\x2 & 2.506 & 2.572 &  & 2.673 & 2.680 &  \\
\fmt 128\x2 & 2.480 & 2.545 &  & 2.644 & 2.651 &  \\
\fmt 512\x4 & 2.386 & 2.447 &  & 2.551 & 2.558 &  \\
\fmt 768\x4 & 2.382 & 2.442 &  & 2.546 & 2.553 &  \\
JPEG XL e9 & 2.626 & 2.691 & 2.63 & 2.787 & 2.794 & 2.79 \\
HTJ2K & 3.145 & 3.217 &  & -- & -- &  \\
JPEG 2000 & 2.932 & 3.000 & 2.93 & -- & -- & 3.12 \\
\bottomrule
\end{tabular}

%% file: supp/tab_paired.tex
\begin{tabular}{@{}lllrrc@{}}
\toprule
Set & \fmt & against & size & 95\% interval & smaller \\
\midrule
Kodak & 16\x2 & JPEG XL e9 & $+$2.6\% & [$+$1.6, $+$3.7]\% & 3/24 \\
 & 64\x2 & JPEG XL e9 & $-$2.3\% & [$-$3.1, $-$1.5]\% & 20/24 \\
 & 64\x2 & JPEG XL e10 & $-$1.7\% & [$-$2.5, $-$0.8]\% & 18/24 \\
 & 64\x2 & FNLIC & $-$2.2\% & [$-$3.2, $-$1.1]\% & 20/24 \\
 & 64\x2 & LC-FDNet & $-$4.1\% & [$-$4.5, $-$3.7]\% & 24/24 \\
 & 64\x2$^\circ$ & LC-FDNet & $+$0.8\% & [$+$0.3, $+$1.2]\% & 7/24 \\
 & 512\x4 & FNLIC & $-$10.2\% & [$-$11.5, $-$8.8]\% & 24/24 \\
 & 512\x4$^\circ$ & LC-FDNet & $-$5.6\% & [$-$6.1, $-$5.1]\% & 24/24 \\
 & 512\x4$^\circ$ & ArIB-BPS & $-$0.2\% & [$-$0.8, $+$0.4]\% & 15/24 \\
 & 768\x4 & 512\x4 & $+$0.1\% & [0.0, $+$0.1]\% & 5/24 \\
\midrule
DIV2K/4 & 16\x2 & JPEG XL e9 & $-$0.1\% & [$-$0.8, $+$0.6]\% & 16/25 \\
 & 64\x2 & JPEG XL e9 & $-$3.8\% & [$-$4.4, $-$3.3]\% & 25/25 \\
 & 64\x2 & FNLIC & $-$2.7\% & [$-$4.4, $-$1.0]\% & 19/25 \\
 & 64\x2 & LC-FDNet & $-$0.7\% & [$-$1.1, $-$0.2]\% & 18/25 \\
 & 512\x4 & FNLIC & $-$7.5\% & [$-$9.5, $-$5.5]\% & 24/25 \\
 & 512\x4 & ArIB-BPS & $+$1.2\% & [$-$0.8, $+$2.6]\% & 6/25 \\
 & 768\x4 & ArIB-BPS & $+$1.0\% & [$-$0.9, $+$2.4]\% & 6/25 \\
 & 768\x4 & 512\x4 & $-$0.2\% & [$-$0.2, $-$0.1]\% & 25/25 \\
\bottomrule
\end{tabular}

%% file: supp/tab_a100.tex
\begin{tabular}{@{}lrrrr@{}}
\toprule
 & \multicolumn{2}{c}{Kodak} & \multicolumn{2}{c}{DIV2K / 4} \\
Codec & bpsp & ms & bpsp & ms \\
\midrule
\fmt fast & 3.981 & 0.22 & 4.145 & 0.93 \\
\fmt dense & 3.187 & 0.70 & 3.072 & 1.4 \\
\fmt 16\x2 & 2.960 & 7.6 & 2.782 & 14.8 \\
\fmt 32\x2 & 2.903 & 8.4 & 2.731 & 19.8 \\
\fmt 64\x2 & 2.818 & 12.6 & 2.678 & 37.5 \\
\fmt 128\x2 & 2.780 & 33.6 & -- & -- \\
\fmt 512\x4 & 2.588 & 58.7 & 2.545 & 245 \\
\fmt 768\x4 & 2.590 & 75.6 & 2.541 & 366 \\
\addlinespace[2pt]
HTJ2K & 3.510 & 0.90 & 3.354 & 2.1 \\
JPEG 2000 & 3.191 & 12.2 & 3.132 & 16.7 \\
LC-FDNet & 2.938 & 997 & 2.696 & 7301 \\
FNLIC & 2.881 & 535 & 2.752 & 3387 \\
ArIB-BPS & 2.781 & 3586 & 2.516 & 25927 \\
\bottomrule
\end{tabular}

%% file: supp/tab_large.tex
\begin{tabular}{@{}lrrrrrrrr@{}}
\toprule
Network & pixels & MP & tables & total & B/px & first & later & MP/s \\
\midrule
16\x2 & 3840\x2160 & 8.3 & 1.8 & 2.1 & 276 & 42 & 33 & 248 \\
 & 5760\x3240 & 18.7 & 4.0 & 4.8 & 276 & 130 & 76 & 246 \\
 & 7680\x4320 & 33.2 & 7.0 & 8.5 & 276 & 257 & 133 & 249 \\
\addlinespace[2pt]
64\x2 & 3840\x2160 & 8.3 & 1.8 & 2.1 & 276 & 109 & 100 & 83 \\
 & 5760\x3240 & 18.7 & 4.0 & 4.8 & 276 & 337 & 221 & 85 \\
 & 7680\x4320 & 33.2 & 7.0 & 8.5 & 276 & 416 & 386 & 86 \\
\addlinespace[2pt]
512\x4 & 3840\x2160 & 8.3 & 1.8 & 3.1 & 403 & 1455 & 1439 & 5.8 \\
 & 5760\x3240 & 18.7 & 4.0 & 5.8 & 335 & 3322 & 3227 & 5.8 \\
 & 7680\x4320 & 33.2 & 7.0 & 9.6 & 311 & 6916 & 7107 & 4.7 \\
\bottomrule
\end{tabular}

%% file: supp/tab_cold.tex
\begin{tabular}{@{}lrrrrrrrr@{}}
\toprule
 & \multicolumn{2}{c}{tables} & \multicolumn{2}{c}{16\x2} & \multicolumn{2}{c}{64\x2} & \multicolumn{2}{c}{512\x4} \\
\cmidrule(lr){2-3}\cmidrule(lr){4-5}\cmidrule(lr){6-7}\cmidrule(l){8-9}
Size & host & GPU & first & later & first & later & first & later \\
\midrule
64\x64 & 0.78 & 0.35 &  &  &  &  &  &  \\
256\x256 & 2.3 & 0.43 &  &  &  &  &  &  \\
768\x512 & 26.3 & 3.4 & 7.2 & 6.1 & 13.3 & 12.5 & 85.5 & 84.8 \\
2040\x1356 & 123 & 7.2 & 19.6 & 13.9 & 45.7 & 39.9 & 500 & 494 \\
3840\x2160 & 370 & 17.1 &  &  &  &  &  &  \\
7680\x4320 & 1782 & 68.8 &  &  &  &  &  &  \\
\bottomrule
\end{tabular}

%% file: supp/tab_coco.tex
\begin{tabular}{@{}lrrrr@{}}
\toprule
 & & \multicolumn{3}{c}{ms per photo, tables of} \\
\cmidrule(l){3-5}
Codec & bpsp & all sizes & 16 sizes & none \\
\midrule
\fmt dense, graph & 3.607 & 1.42 & 1.48 & -- \\
\fmt dense, eager &  & -- & 0.94 & 0.80 \\
\fmt 16\x2 & 3.458 & 6.21 & 7.20 & 7.69 \\
\fmt 64\x2 & 3.311 & 11.0 & 12.4 & 12.7 \\
JPEG XL e9 & 3.274 &  &  &  \\
WebP & 3.563 &  &  &  \\
\bottomrule
\end{tabular}

%% file: supp/tab_host.tex
\begin{tabular}{@{}llrrrrr@{}}
\toprule
Set & Codec & bpsp & ms & GPU & share & traced \\
\midrule
Kodak & HTJ2K & 3.510 & 0.73 & 0.60 & 82\% & 1.73 \\
 & dense, $T$=512 & 3.187 & 0.54 & 0.43 & 79\% & 0.59 \\
 & dense, $T$=2,048 & 3.150 & 1.17 & 1.05 & 90\% & 1.23 \\
\addlinespace[2pt]
CLIC.p/4 & HTJ2K & 2.960 & 1.35 & 0.99 & 73\% & 2.13 \\
 & dense, $T$=512 & 2.671 & 1.18 & 0.95 & 81\% & 1.22 \\
 & dense, $T$=2,048 & 2.635 & 1.62 & 1.37 & 85\% & 1.64 \\
\addlinespace[2pt]
DIV2K/4 & HTJ2K & 3.354 & 1.48 & 1.10 & 75\% & 2.23 \\
 & dense, $T$=512 & 3.071 & 1.36 & 1.09 & 81\% & 1.41 \\
 & dense, $T$=2,048 & 3.034 & 1.75 & 1.48 & 85\% & 1.79 \\
\bottomrule
\end{tabular}

%% file: supp/tab_nvcomp.tex
\begin{tabular}{@{}llrr@{}}
\toprule
Set & nvCOMP variant & bpsp & MP/s \\
\midrule
Kodak & Deflate, Sub, 64k & 4.535 & 998 \\
 & GDeflate, Sub, 64k & 4.559 & 1485 \\
 & GDeflate, Sub, 16k & 4.791 & 1952 \\
 & ANS, Sub, 64k & 5.459 & 2026 \\
 & GDeflate, raw, 16k & 5.596 & 2042 \\
 & Bitcomp, raw, 64k & 6.904 & 2205 \\
\addlinespace[2pt]
CLIC.p / 4 & Deflate, Sub, 64k & 3.827 & 2096 \\
 & GDeflate, Sub, 64k & 3.850 & 2671 \\
 & GDeflate, Sub, 16k & 4.030 & 2721 \\
 & ANS, Sub, 64k & 4.545 & 2893 \\
 & GDeflate, raw, 64k & 5.482 & 3364 \\
\addlinespace[2pt]
DIV2K / 4 & Deflate, Sub, 64k & 4.624 & 2193 \\
 & GDeflate, Sub, 64k & 4.648 & 2799 \\
 & ANS, Sub, 64k & 5.422 & 3060 \\
 & Bitcomp, raw, 64k & 6.696 & 3139 \\
\bottomrule
\end{tabular}

%% file: refs.bib
@misc{hpac,
  author        = {Li, Daxin and Bai, Yuanchao and Wang, Kai and Zhao, Wenbo and Jiang, Junjun and Liu, Xianming},
  title         = {Rethinking Autoregressive Models for Lossless Image Compression via Hierarchical Parallelism and Progressive Adaptation},
  howpublished  = {arXiv preprint arXiv:2511.10991},
  eprint        = {2511.10991},
  archivePrefix = {arXiv},
  year          = {2025},
}

@inproceedings{callic,
  author    = {Li, Daxin and Bai, Yuanchao and Wang, Kai and Jiang, Junjun and Liu, Xianming and Gao, Wen},
  title     = {{CALLIC}: Content Adaptive Learning for Lossless Image Compression},
  booktitle = {AAAI},
  volume    = {39},
  pages     = {4679--4688},
  year      = {2025},
  doi       = {10.1609/aaai.v39i5.32494},
}

@article{dlpr,
  author  = {Bai, Yuanchao and Liu, Xianming and Wang, Kai and Ji, Xiangyang and Wu, Xiaolin and Gao, Wen},
  title   = {Deep Lossy Plus Residual Coding for Lossless and Near-Lossless Image Compression},
  journal = {IEEE TPAMI},
  volume  = {46},
  number  = {5},
  pages   = {3577--3594},
  year    = {2024},
  doi     = {10.1109/TPAMI.2023.3348486},
}

@inproceedings{aribbps,
  author    = {Zhang, Zhe and Wang, Huairui and Chen, Zhenzhong and Liu, Shan},
  title     = {Learned Lossless Image Compression based on Bit Plane Slicing},
  booktitle = {CVPR},
  pages     = {27569--27578},
  year      = {2024},
}

@inproceedings{lcfdnet,
  author    = {Rhee, Hochang and Jang, Yeong Il and Kim, Seyun and Cho, Nam Ik},
  title     = {{LC-FDNet}: Learned Lossless Image Compression With Frequency Decomposition Network},
  booktitle = {CVPR},
  pages     = {6023--6032},
  year      = {2022},
}

@inproceedings{fnlic,
  author    = {Zhang, Zhe and Chen, Zhenzhong and Liu, Shan},
  title     = {Fitted Neural Lossless Image Compression},
  booktitle = {CVPR},
  pages     = {23249--23258},
  year      = {2025},
}

@inproceedings{l3c,
  author    = {Mentzer, Fabian and Agustsson, Eirikur and Tschannen, Michael and Timofte, Radu and {Van Gool}, Luc},
  title     = {Practical Full Resolution Learned Lossless Image Compression},
  booktitle = {CVPR},
  pages     = {10629--10638},
  year      = {2019},
  doi       = {10.1109/CVPR.2019.01088},
}

@inproceedings{rc,
  author    = {Mentzer, Fabian and {Van Gool}, Luc and Tschannen, Michael},
  title     = {Learning Better Lossless Compression Using Lossy Compression},
  booktitle = {CVPR},
  pages     = {6637--6646},
  year      = {2020},
}

@inproceedings{idf,
  author    = {Hoogeboom, Emiel and Peters, Jorn and {van den Berg}, Rianne and Welling, Max},
  title     = {Integer Discrete Flows and Lossless Compression},
  booktitle = {NeurIPS},
  volume    = {32},
  pages     = {12134--12144},
  year      = {2019},
}

@inproceedings{ivpf,
  author    = {Zhang, Shifeng and Zhang, Chen and Kang, Ning and Li, Zhenguo},
  title     = {{iVPF}: Numerical Invertible Volume Preserving Flow for Efficient Lossless Compression},
  booktitle = {CVPR},
  pages     = {620--629},
  year      = {2021},
  doi       = {10.1109/CVPR46437.2021.00068},
}

@inproceedings{iflow,
  author    = {Zhang, Shifeng and Kang, Ning and Ryder, Tom and Li, Zhenguo},
  title     = {{iFlow}: Numerically Invertible Flows for Efficient Lossless Compression via a Uniform Coder},
  booktitle = {NeurIPS},
  volume    = {34},
  pages     = {5822--5833},
  year      = {2021},
}

@misc{srec,
  author        = {Cao, Sheng and Wu, Chao-Yuan and Kr{\"a}henb{\"u}hl, Philipp},
  title         = {Lossless Image Compression through Super-Resolution},
  howpublished  = {arXiv preprint arXiv:2004.02872},
  eprint        = {2004.02872},
  archivePrefix = {arXiv},
  year          = {2020},
}

@inproceedings{pixelcnnpp,
  author    = {Salimans, Tim and Karpathy, Andrej and Chen, Xi and Kingma, Diederik P.},
  title     = {{PixelCNN++}: Improving the {PixelCNN} with Discretized Logistic Mixture Likelihood and Other Modifications},
  booktitle = {ICLR},
  year      = {2017},
}

@inproceedings{pixelcnn,
  author    = {{van den Oord}, A{\"a}ron and Kalchbrenner, Nal and Kavukcuoglu, Koray},
  title     = {Pixel Recurrent Neural Networks},
  booktitle = {ICML},
  series    = {PMLR},
  volume    = {48},
  pages     = {1747--1756},
  year      = {2016},
}

@inproceedings{multiscale,
  author    = {Reed, Scott and {van den Oord}, A{\"a}ron and Kalchbrenner, Nal and {G{\'o}mez Colmenarejo}, Sergio and Wang, Ziyu and Chen, Yutian and Belov, Dan and {de Freitas}, Nando},
  title     = {Parallel Multiscale Autoregressive Density Estimation},
  booktitle = {ICML},
  series    = {PMLR},
  volume    = {70},
  pages     = {2912--2921},
  year      = {2017},
}

@inproceedings{checkerboard,
  author    = {He, Dailan and Zheng, Yaoyan and Sun, Baocheng and Wang, Yan and Qin, Hongwei},
  title     = {Checkerboard Context Model for Efficient Learned Image Compression},
  booktitle = {CVPR},
  pages     = {14771--14780},
  year      = {2021},
}

@inproceedings{elic,
  author    = {He, Dailan and Yang, Ziming and Peng, Weikun and Ma, Rui and Qin, Hongwei and Wang, Yan},
  title     = {{ELIC}: Efficient Learned Image Compression With Unevenly Grouped Space-Channel Contextual Adaptive Coding},
  booktitle = {CVPR},
  pages     = {5708--5717},
  year      = {2022},
}

@inproceedings{jpegxl,
  author    = {Alakuijala, Jyrki and {van Asseldonk}, Ruud and Boukortt, Sami and Bruse, Martin and Comsa, Iulia-Maria and Firsching, Moritz and Fischbacher, Thomas and Gomez, Sebastian and Kliuchnikov, Evgenii and Obryk, Robert and Potempa, Krzysztof and Rhatushnyak, Alexander and Sneyers, Jon and Szabadka, Zoltan and Vandevenne, Lode and Versari, Luca and Wassenberg, Jan},
  title     = {{JPEG XL} Next-Generation Image Compression Architecture and Coding Tools},
  booktitle = {Applications of Digital Image Processing XLII},
  series    = {Proc. SPIE},
  volume    = {11137},
  pages     = {111370K},
  year      = {2019},
  doi       = {10.1117/12.2529237},
}

@inproceedings{htj2k,
  author    = {Taubman, David and Naman, Aous Thabit and Mathew, Reji},
  title     = {High Throughput Block Coding in the {HTJ2K} Compression Standard},
  booktitle = {ICIP},
  pages     = {1079--1083},
  year      = {2019},
}

@book{jpeg2000,
  author    = {Taubman, David S. and Marcellin, Michael W.},
  title     = {{JPEG2000}: Image Compression Fundamentals, Standards and Practice},
  publisher = {Kluwer Academic Publishers},
  series    = {The Springer International Series in Engineering and Computer Science},
  volume    = {642},
  year      = {2002},
  doi       = {10.1007/978-1-4615-0799-4},
}

@article{locoi,
  author  = {Weinberger, Marcelo J. and Seroussi, Gadiel and Sapiro, Guillermo},
  title   = {The {LOCO-I} Lossless Image Compression Algorithm: Principles and Standardization into {JPEG-LS}},
  journal = {IEEE TIP},
  volume  = {9},
  number  = {8},
  pages   = {1309--1324},
  year    = {2000},
  doi     = {10.1109/83.855427},
}

@article{calic,
  author  = {Wu, Xiaolin and Memon, Nasir},
  title   = {Context-Based, Adaptive, Lossless Image Coding},
  journal = {IEEE TCOM},
  volume  = {45},
  number  = {4},
  pages   = {437--444},
  year    = {1997},
  doi     = {10.1109/26.585919},
}

@inproceedings{flif,
  author    = {Sneyers, Jon and Wuille, Pieter},
  title     = {{FLIF}: Free Lossless Image Format Based on {MANIAC} Compression},
  booktitle = {ICIP},
  pages     = {66--70},
  year      = {2016},
}

@article{hint,
  author  = {Roos, P. and Viergever, M. A. and {van Dijke}, M. C. A. and Peters, J. H.},
  title   = {Reversible Intraframe Compression of Medical Images},
  journal = {IEEE TMI},
  volume  = {7},
  number  = {4},
  pages   = {328--336},
  year    = {1988},
  doi     = {10.1109/42.14516},
}

@article{sp,
  author  = {Said, Amir and Pearlman, William A.},
  title   = {An Image Multiresolution Representation for Lossless and Lossy Compression},
  journal = {IEEE TIP},
  volume  = {5},
  number  = {9},
  pages   = {1303--1310},
  year    = {1996},
}

@article{lifting,
  author  = {Calderbank, A. R. and Daubechies, Ingrid and Sweldens, Wim and Yeo, Boon-Lock},
  title   = {Wavelet Transforms That Map Integers to Integers},
  journal = {Appl. Comput. Harmon. Anal.},
  volume  = {5},
  number  = {3},
  pages   = {332--369},
  year    = {1998},
  doi     = {10.1006/acha.1997.0238},
}

@misc{ans,
  author        = {Duda, Jarek},
  title         = {Asymmetric Numeral Systems: Entropy Coding Combining Speed of {Huffman} Coding with Compression Rate of Arithmetic Coding},
  howpublished  = {arXiv preprint arXiv:1311.2540},
  eprint        = {1311.2540},
  archivePrefix = {arXiv},
  year          = {2013},
}

@misc{interleaved,
  author        = {Giesen, Fabian},
  title         = {Interleaved Entropy Coders},
  howpublished  = {arXiv preprint arXiv:1402.3392},
  eprint        = {1402.3392},
  archivePrefix = {arXiv},
  year          = {2014},
}

@inproceedings{gpuans,
  author    = {Wei{\ss}enberger, Andr{\'e} and Schmidt, Bertil},
  title     = {Massively Parallel {ANS} Decoding on {GPUs}},
  booktitle = {ICPP},
  pages     = {100:1--100:10},
  year      = {2019},
  doi       = {10.1145/3337821.3337888},
}

@inproceedings{gpuhuff,
  author    = {Wei{\ss}enberger, Andr{\'e} and Schmidt, Bertil},
  title     = {Massively Parallel {Huffman} Decoding on {GPUs}},
  booktitle = {ICPP},
  pages     = {27:1--27:10},
  year      = {2018},
  doi       = {10.1145/3225058.3225076},
}

@inproceedings{recoil,
  author    = {Lin, Fangzheng and Arunruangsirilert, Kasidis and Sun, Heming and Katto, Jiro},
  title     = {Recoil: Parallel {rANS} Decoding with Decoder-Adaptive Scalability},
  booktitle = {ICPP},
  pages     = {31--40},
  year      = {2023},
  doi       = {10.1145/3605573.3605588},
}

@misc{kodak,
  author       = {{Eastman Kodak Company}},
  title        = {Kodak Lossless True Color Image Suite},
  howpublished = {\url{http://r0k.us/graphics/kodak/}},
  note         = {24 images from Kodak PhotoCD PCD0992},
}

@inproceedings{div2k,
  author    = {Agustsson, Eirikur and Timofte, Radu},
  title     = {{NTIRE} 2017 Challenge on Single Image Super-Resolution: Dataset and Study},
  booktitle = {CVPR Workshops},
  pages     = {1122--1131},
  year      = {2017},
}

@misc{clic,
  author       = {Toderici, George and Shi, Wenzhe and Timofte, Radu and Theis, Lucas and Ball{\'e}, Johannes and Agustsson, Eirikur and Johnston, Nick and Mentzer, Fabian},
  title        = {Workshop and Challenge on Learned Image Compression ({CLIC} 2020)},
  howpublished = {CVPR 2020 Workshops, \url{http://www.compression.cc}},
  year         = {2020},
}

@inproceedings{flickr2k,
  author    = {Lim, Bee and Son, Sanghyun and Kim, Heewon and Nah, Seungjun and Lee, Kyoung Mu},
  title     = {Enhanced Deep Residual Networks for Single Image Super-Resolution},
  booktitle = {CVPR Workshops},
  pages     = {1132--1140},
  year      = {2017},
}

@misc{nvjpeg2000,
  author       = {{NVIDIA}},
  title        = {{nvJPEG2000}: {GPU}-Accelerated {JPEG} 2000 Decoding and Encoding Library},
  howpublished = {\url{https://developer.nvidia.com/nvjpeg}},
  note         = {Documentation: \url{https://docs.nvidia.com/cuda/nvjpeg2000/}. Accessed 2026-11-03},
  year          = {2026},
}

@misc{nvcomp,
  author       = {{NVIDIA}},
  title        = {{nvCOMP}: {GPU}-Accelerated Data Compression and Decompression Library},
  howpublished = {\url{https://developer.nvidia.com/nvcomp}},
  note         = {Accessed 2026-11-03},
  year          = {2026},
}

@misc{nvimagecodec,
  author       = {{NVIDIA}},
  title        = {{nvImageCodec}: {GPU}- and {CPU}-Accelerated Image Codecs with a Unified Interface},
  howpublished = {\url{https://github.com/NVIDIA/nvImageCodec}},
  note         = {Accessed 2026-11-03},
  year          = {2026},
}

@misc{png,
  author       = {Duce, David},
  title        = {Portable Network Graphics ({PNG}) Specification (Second Edition)},
  howpublished = {W3C Recommendation, \url{https://www.w3.org/TR/2003/REC-PNG-20031110/}},
  note         = {Also published as ISO/IEC 15948:2003},
  month        = nov,
  year         = {2003},
}

@misc{webp,
  author       = {Zern, James and Massimino, Pascal and Alakuijala, Jyrki},
  title        = {{WebP} Image Format},
  howpublished = {RFC 9649, IETF, \url{https://www.rfc-editor.org/rfc/rfc9649}},
  month        = nov,
  year         = {2024},
  doi          = {10.17487/RFC9649},
}

@misc{qoi,
  author       = {Szablewski, Dominic},
  title        = {{QOI}: The Quite {OK} Image Format},
  howpublished = {\url{https://qoiformat.org}},
  note         = {Specification version 1.0 dated 2022-01-05},
  year         = {2021},
}

@inproceedings{igpt,
  author    = {Chen, Mark and Radford, Alec and Child, Rewon and Wu, Jeffrey and Jun, Heewoo and Luan, David and Sutskever, Ilya},
  title     = {Generative Pretraining From Pixels},
  booktitle = {ICML},
  series    = {PMLR},
  volume    = {119},
  pages     = {1691--1703},
  year      = {2020},
}

@misc{statetrees,
  author        = {Ali, Ammar and Lefkimmiatis, Stamatis},
  title         = {Every Weight Is a Path of Shared Discrete Decisions},
  howpublished  = {arXiv preprint arXiv:2610.31931},
  eprint        = {2610.31931},
  archivePrefix = {arXiv},
  year          = {2026},
}

@article{llmvp,
  title   = {Large Language Model for Lossless Image Compression with Visual Prompts},
  author  = {Du, Junhao and Zhou, Chuqin and Cao, Ning and Chen, Gang and Chen, Yunuo and Cheng, Zhengxue and Song, Li and Lu, Guo and Zhang, Wenjun},
  journal = {arXiv preprint arXiv:2502.16163},
  year    = {2025}
}

@misc{ffv1,
  author       = {Niedermayer, Michael and Rice, Dave and Martinez, J{\'e}r{\^o}me},
  title        = {{FFV1} Video Coding Format Versions 0, 1, and 3},
  howpublished = {RFC 9043, Internet Engineering Task Force},
  year         = {2021}
}

@article{kamisli,
  title   = {Learned Lossless Image Compression Through Interpolation With Low Complexity},
  author  = {Kamisli, Fatih},
  journal = {IEEE Transactions on Circuits and Systems for Video Technology},
  volume  = {33},
  number  = {12},
  pages   = {7832--7841},
  year    = {2023}
}

@inproceedings{dlic,
  author    = {Barzen, Benjamin Lukas Cajus and Glazov, Fedor and Geistert, Jonas and Sikora, Thomas},
  title     = {Accelerated Deep Lossless Image Coding with Unified Paralleleized {GPU} Coding Architecture},
  booktitle = {PCS},
  pages     = {109--113},
  year      = {2022},
}

@article{gumus,
  author  = {G{\"u}m{\"u}{\c{s}}, Sinem and Kam{\i}{\c{s}}l{\i}, Fatih},
  title   = {A Learned Pixel-by-Pixel Lossless Image Compression Method with {59K} Parameters and Parallel Decoding},
  journal = {Multimedia Tools and Applications},
  volume  = {83},
  number  = {8},
  pages   = {22975--22993},
  year    = {2024},
  doi     = {10.1007/s11042-023-16270-4},
}

@inproceedings{pilc,
  title     = {{PILC}: Practical Image Lossless Compression with an End-to-End {GPU} Oriented Neural Framework},
  author    = {Kang, Ning and Qiu, Shanzhao and Zhang, Shifeng and Li, Zhenguo and Xia, Shu-Tao},
  booktitle = {CVPR},
  pages     = {3729--3738},
  year      = {2022}
}

@article{kimcho,
  title   = {Hierarchical Prediction and Context Adaptive Coding for Lossless Color Image Compression},
  author  = {Kim, Seyun and Cho, Nam Ik},
  journal = {IEEE Transactions on Image Processing},
  volume  = {23},
  number  = {1},
  pages   = {445--449},
  year    = {2014}
}

@inproceedings{mspsm,
  title     = {Lossless Image Compression Using a Multi-scale Progressive Statistical Model},
  author    = {Zhang, Honglei and Cricri, Francesco and Tavakoli, Hamed R. and Zou, Nannan and Aksu, Emre and Hannuksela, Miska M.},
  booktitle = {ACCV},
  pages     = {609--622},
  year      = {2020}
}

@inproceedings{spn,
  title     = {Generating High Fidelity Images with Subscale Pixel Networks and Multidimensional Upscaling},
  author    = {Menick, Jacob and Kalchbrenner, Nal},
  booktitle = {ICLR},
  year      = {2019}
}

@inproceedings{iodf,
  title     = {Fast Lossless Neural Compression with Integer-Only Discrete Flows},
  author    = {Wang, Siyu and Chen, Jianfei and Li, Chongxuan and Zhu, Jun and Zhang, Bo},
  booktitle = {ICML},
  pages     = {22562--22575},
  year      = {2022}
}

@inproceedings{intnets,
  title     = {Integer Networks for Data Compression with Latent-Variable Models},
  author    = {Ball{\'e}, Johannes and Johnston, Nick and Minnen, David},
  booktitle = {ICLR},
  year      = {2019}
}

@inproceedings{rozendaal,
  title     = {Overfitting for Fun and Profit: Instance-Adaptive Data Compression},
  author    = {van Rozendaal, Ties and Huijben, Iris A. M. and Cohen, Taco S.},
  booktitle = {ICLR},
  year      = {2021}
}

@article{jxloverview,
  title   = {The {JPEG XL} Image Coding System: History, Features, Coding Tools, Design Rationale, and Future},
  author  = {Sneyers, Jon and Alakuijala, Jyrki and Versari, Luca and Szabadka, Zolt{\'a}n and Boukortt, Sami and Cohen-Tidhar, Amnon and Firsching, Moritz and Kliuchnikov, Evgenii and Lev-Ami, Tal and Portis, Eric and Richter, Thomas and Watanabe, Osamu},
  journal = {arXiv preprint arXiv:2506.05987},
  year    = {2025}
}

@misc{cublas,
  author       = {{NVIDIA}},
  title        = {{cuBLAS} Library User Guide},
  howpublished = {\url{https://docs.nvidia.com/cuda/cublas/}},
  note         = {Section ``Results reproducibility''. Accessed 2026-11-04},
  year         = {2026},
}

@inproceedings{hilloc,
  author    = {James Townsend and Thomas Bird and Julius Kunze and David Barber},
  title     = {{HiLLoC}: Lossless Image Compression with Hierarchical Latent Variable Models},
  booktitle = {ICLR},
  year      = {2020},
}

@misc{dietgpu,
  author       = {Jeff Johnson},
  title        = {{DietGPU}: {GPU}-based lossless compression for numerical data},
  howpublished = {\url{https://github.com/facebookresearch/dietgpu}},
  note         = {Accessed 2026-11-04},
  year         = {2022},
}

@misc{gralic,
  author       = {Alex Rhatushnyak},
  title        = {{GraLIC}: lossless image compressor, version 1.11},
  year         = {2011},
  howpublished = {Lossless Photo Compression Benchmark, \url{https://imagecompression.info/gralic/LPCB.html}}
}

@article{expgolomb,
  author  = {Jukka Teuhola},
  title   = {A compression method for clustered bit-vectors},
  journal = {Information Processing Letters},
  volume  = {7},
  number  = {6},
  pages   = {308--311},
  year    = {1978}
}

@article{jpeg,
  author  = {Gregory K. Wallace},
  title   = {The {JPEG} still picture compression standard},
  journal = {IEEE Transactions on Consumer Electronics},
  volume  = {38},
  number  = {1},
  pages   = {xviii--xxxiv},
  year    = {1992}
}

@inproceedings{shvc,
  author    = {Ryder, Tom and Zhang, Chen and Kang, Ning and Zhang, Shifeng},
  title     = {Split Hierarchical Variational Compression},
  booktitle = {CVPR},
  pages     = {386--395},
  year      = {2022},
}

@inproceedings{pcircuits,
  author    = {Liu, Anji and Mandt, Stephan and {Van den Broeck}, Guy},
  title     = {Lossless Compression with Probabilistic Circuits},
  booktitle = {ICLR},
  year      = {2022},
}

@inproceedings{dcvcrt,
  author    = {Jia, Zhaoyang and Li, Bin and Li, Jiahao and Xie, Wenxuan and Qi, Linfeng and Li, Houqiang and Lu, Yan},
  title     = {Towards Practical Real-Time Neural Video Compression},
  booktitle = {CVPR},
  pages     = {12543--12552},
  year      = {2025},
}

@inproceedings{coolchic,
  title     = {{COOL-CHIC}: Coordinate-based Low Complexity Hierarchical Image Codec},
  author    = {Ladune, Th{\'e}o and Philippe, Pierrick and Henry, F{\'e}lix and Clare, Gordon and Leguay, Thomas},
  booktitle = {ICCV},
  year      = {2023}
}

@inproceedings{c3,
  title     = {{C3}: High-Performance and Low-Complexity Neural Compression from a Single Image or Video},
  author    = {Kim, Hyunjik and Bauer, Matthias and Theis, Lucas and Schwarz, Jonathan Richard and Dupont, Emilien},
  booktitle = {CVPR},
  year      = {2024}
}

@inproceedings{film,
  title     = {{FiLM}: Visual Reasoning with a General Conditioning Layer},
  author    = {Perez, Ethan and Strub, Florian and de Vries, Harm and Dumoulin, Vincent and Courville, Aaron},
  booktitle = {AAAI},
  year      = {2018}
}

@inproceedings{lora,
  title     = {{LoRA}: Low-Rank Adaptation of Large Language Models},
  author    = {Hu, Edward J. and Shen, Yelong and Wallis, Phillip and Allen-Zhu, Zeyuan and Li, Yuanzhi and Wang, Shean and Wang, Lu and Chen, Weizhu},
  booktitle = {ICLR},
  year      = {2022}
}
